\documentclass{article}
\usepackage{iclr2027_conference,times}

\usepackage{amsmath,amsfonts,bm}

\def\eqref#1{equation~\ref{#1}}

\def\1{\bm{1}}

\DeclareMathAlphabet{\mathsfit}{\encodingdefault}{\sfdefault}{m}{sl}
\SetMathAlphabet{\mathsfit}{bold}{\encodingdefault}{\sfdefault}{bx}{n}

\usepackage{hyperref}
\usepackage{url}
\let\cite\citep

\usepackage{graphicx}
\usepackage{amsmath}
\usepackage{amssymb}
\usepackage{amsthm}
\usepackage{enumitem}
\usepackage{colortbl}
\usepackage{array}
\usepackage{multirow}
\usepackage{longtable}
\usepackage{makecell}
\usepackage{wrapfig}
\usepackage{tabularray}
\usepackage{pifont}
\usepackage{subcaption}

\usepackage{float}
\usepackage{tikz}
\usepackage{tcolorbox}
\tcbuselibrary{skins,breakable}
\usepackage{bibunits}
\usepackage{setspace}
\usepackage{booktabs}
\usepackage{xcolor}

\definecolor{cvprblue}{rgb}{0.21, 0.49, 0.74}
\hypersetup{
    colorlinks=true,
    linkcolor=cvprblue,
    citecolor=cvprblue,
    urlcolor=cvprblue
}

\usepackage{placeins}
\usepackage[ruled]{algorithm2e}
\usepackage[normalem]{ulem}
\SetKwInput{KwIn}{Input}
\SetKwInput{KwOut}{Output}
\SetKw{KwRet}{Return}

\title{Towards Unbiased On-Policy Distillation for Block Diffusion Language Models}

\author{
\begin{minipage}[t]{0.96\textwidth}
    \centering
    Zaiquan Yang$^{1,2,*}$ \quad Fei Wei$^{2,\dagger}$ \quad Yong Wang$^{2,\dagger}$ \quad Yudong Han$^{2}$ \quad Yiyu Li$^{1}$\\
    Zhuofan Zong$^{3}$ \quad Gerhard Petrus Hancke$^{1}$ \quad Xiangxiang Chu$^{2}$ \quad Rynson WH Lau$^{4}$\\[0.6em]
    {\normalfont\small
        $^{1}$\,City University of Hong Kong \quad $^{2}$\,AMAP, Alibaba Group\\
        $^{3}$\,The Chinese University of Hong Kong\\
        $^{4}$\,City University of Hong Kong (Dongguan)
    }
\end{minipage}
}

\newcommand{\ryn}[1]{\textcolor{black}{#1}}
\newcommand{\rynq}[1]{\textcolor{black}{#1}}

\theoremstyle{plain}
\newtheorem{proposition}{Proposition}

\newif\ifincludeappendix
\includeappendixtrue

\ifincludeappendix\else

\fi

\iclrfinalcopy
\begin{document}

\maketitle
\pagestyle{plain}
\thispagestyle{plain}

\begingroup
\renewcommand{\thefootnote}{\fnsymbol{footnote}}
\footnotetext[1]{Work done during the internship at AMAP, Alibaba Group.}
\footnotetext[2]{Project lead.}
\endgroup

\begin{abstract}
On-policy distillation (OPD) has emerged as an effective post-training paradigm for language models, with recent efforts extending it to block diffusion language models (BDLMs). However, existing studies focus almost exclusively on small block sizes, leaving distillation into student models with larger blocks underexplored.
In this work, we investigate this regime and reveal two critical optimization biases that induce severe training instability.
First, mismatched block boundaries between teacher and student cause \textbf{\textit{context misalignment}}, providing distorted supervisory signals that misguide student decoding.
Second, even under aligned contexts, an \textbf{\textit{intrinsic optimization bias}} in OPD, where the student tends to rapidly absorb high-support signals while lagging on low-support updates, drives a premature confidence surge that traps weaker students in catastrophic overconfidence collapse.
To resolve these problems, we propose \mbox{\textbf{Un-OPD}}, an unbiased on-policy distillation framework with two novelties for stabilizing BDLM training.
First, Un-OPD introduces a boundary-aware step filtering strategy that eliminates context-misaligned decoding steps.
Second, Un-OPD proposes moderating optimization intensity at high-support positions via a support-rebalanced confidence calibration, thereby bypassing overconfidence collapse.
Beyond stability, we also introduce a rollout reuse mechanism to reduce rollout generation overhead.
Extensive experiments on math reasoning and code generation benchmarks show that Un-OPD consistently stabilizes training and delivers superior performance while reducing wall-clock training time by approximately half.
\end{abstract}

\section{Introduction}
\label{sec:intro}

Diffusion large language models (dLLMs) have recently emerged as a compelling alternative to autoregressive (AR) models~\cite{achiam2023gpt,hui2024qwen2,yang2025qwen3} by generating multiple tokens in parallel through iterative denoising~\cite{nie2025large,ye2025dream}.
For example, some open-source models like LLaDA~\cite{nie2025large} and Dream~\cite{ye2025dream} maintain a long sequence of mask tokens and iteratively reveal tokens through parallel denoising steps.
To inherit the advantages of AR models, particularly KV caching compatibility~\cite{kwon2023efficient,ni2026flexibility} and flexible sequence extension, recent dLLM works have advanced toward block diffusion large language models (BDLMs)~\cite{arriola2025block,cheng2026sdar,tian2025next,wu2026dreamreasoner,bie2025llada2,bie2026llada2}.
By decomposing a sequence into consecutive blocks, BDLMs perform global reasoning autoregressively across blocks while generating tokens via parallel diffusion within each block, achieving a superior quality--latency trade-off.

Following the success of post-training in AR models, particularly reinforcement learning with verifiable rewards (RLVR)~\cite{shao2024deepseekmath,yu2026dapo}, reinforcement learning (RL) has been actively adapted to enhance the reasoning capabilities of dLLMs~\cite{pan2026d,zhao2026d1,wang2026revolutionizing,ou2026principled,ma2026consolidating,ni2026flexibility}.
Specifically, to bypass the intractable conditional likelihood estimation inherent in diffusion denoising, pioneering efforts introduce mean-field surrogates~\cite{zhao2026d1,tang2026wd1} or tree-structured policy updates~\cite{pan2026d}.
%
%
Nevertheless, these approaches either rely on coarse outcome-based rewards \ryn{leading} to inaccurate and sparse credit assignment across multi-step denoising trajectories~\cite{shao2026back,zhao2026d1}, or incur substantial inference overhead to acquire finer-grained reward signals~\cite{pan2026d}.
Recently, on-policy distillation (OPD)~\cite{hou2026uni,yang2026learning} has emerged as another promising post-training paradigm. By aligning the student with teacher predictions along its own rollout trajectories, OPD provides dense, token-level supervision.
Recent studies~\cite{hou2026uni,yang2026learning,fu2026revisiting} demonstrate that OPD outperforms previous RLVR algorithms on language models in terms of both optimization stability and sample efficiency.

Despite extensive progress in AR models, OPD for dLLMs remains at an early stage.
dOPSD~\cite{dat2026dopsd} extends self-distillation to full-sequence denoising models such as \rynq{LLaDA~\cite{nie2025large}} using predictions from later states with more revealed tokens to supervise earlier states. This design is less applicable to BDLMs, where later states reveal context only within the current block.
TOPD~\cite{ren2026trace} extends OPD to BDLMs by distilling along the student's denoising trajectories, but it focuses exclusively on small block sizes, leaving distillation into students with larger blocks, which offer inherently greater parallel decoding potential, largely underexplored.
%
%
In this work, we investigate scaling OPD \ryn{to} BDLMs with larger block sizes.
We use small-block BDLMs as teachers because they provide higher-quality generations, and distill \ryn{them} into students with varying block sizes.

Our work uncovers two underlying optimization biases that impede the effective distillation of BDLMs:
\textbf{(1) Context misalignment.} 
When denoising within a larger block, the student receives biased supervision because the teacher cannot access its authentic decoding states across mismatched boundaries. Such signals tend to be overly aggressive or excessively conservative, disrupting the student's confidence-driven decisions.
\textbf{(2) Biased learning dynamics.} 
During OPD, the student prioritizes tokens with strong teacher support over weakly supported ones, causing confidence to prematurely surge and then fall back. This spurious surge drives fragile students into low-quality decoding regimes that are difficult to rectify.
To address these \ryn{two problems}, we propose \mbox{\textbf{Un-OPD}} (Unbiased On-Policy Distillation), a simple framework for scaling OPD to BDLMs with larger block sizes.
Specifically, we first restrict OPD to decoding states with context alignment, ensuring impartial teacher supervision signals.
We then quantify \textit{teacher support} to distinguish strongly supported positions and rebalance the teacher support by moderating the update intensity, preventing overconfidence collapse.
Beyond stability, we further introduce a \textit{rollout reuse} mechanism that reuses sampled trajectories for reducing rollout generation overhead.
%

In summary, the \ryn{main} contributions {of this work} are as follows:
\begin{itemize}[leftmargin=*, itemsep=4pt, topsep=2pt, parsep=0pt]
    \item We pioneer the study of OPD on BDLMs with varying block sizes, identifying two underlying optimization bottlenecks: \textit{context misalignment} arising from mismatched block sizes
    and \textit{biased learning dynamics} inherent to standard OPD.
    \item  We propose \mbox{\textbf{Un-OPD}}, an unbiased OPD framework that integrates context-aligned step selection and support-rebalanced confidence calibration to stabilize the training dynamics.
    We also introduce a rollout reuse mechanism to reduce rollout overhead during training.
    \item We conduct a comprehensive evaluation across various block sizes and
    demonstrate that Un-OPD achieves superior performance over counterparts, alongside greater training efficiency.
\end{itemize}

\section{Related Work}

\textbf{Block Diffusion Language Models.}
\rynq{dLLMs~\cite{nie2025large,zhu2025llada,ye2025dream}} offer an alternative to
autoregressive LLMs~\cite{yang2025qwen3,achiam2023gpt} through iterative discrete
denoising, with recent gains from scaling data and model size.
\rynq{LLaDA~\cite{nie2025large,zhu2025llada}} trains from scratch via masking and unmasking,
whereas Dream~\cite{ye2025dream} adapts pretrained autoregressive models.
Full-sequence dLLMs, however, struggle to combine coherent chain-of-thought
reasoning with efficient inference mechanisms such as KV caching~\cite{kwon2023efficient,ni2026flexibility}.
Block diffusion language models (BDLMs)~\cite{arriola2025block,cheng2026sdar,tian2025next,wu2026dreamreasoner,bie2025llada2,bie2026llada2}
address this trade-off by generating blocks autoregressively and denoising tokens
in parallel within each block.
Typically adapted from autoregressive models, BDLMs aim to retain their reasoning
capabilities while enabling parallel intra-block generation.

\textbf{On-Policy Distillation.}
By aligning student-generated rollouts with per-token teacher predictions, on-policy distillation (OPD)~\cite{hou2026uni, yang2026learning, gu2026co,fu2026revisiting,xie2026position,xu2026tip,zhang2026fast,jin2026entropy,wang2026trace,xing2026trust} has demonstrated superior post-training efficiency compared to GRPO-style optimization~\cite{shao2024deepseekmath,yu2026dapo,gao2025soft,tan2025gtpo} and off-policy counterparts~\cite{wang2023self,ren2025learning,wu2026generalization}.
To extract more reliable teacher supervision, several works~\cite{zhang2026fast,wang2026trace,xu2026tip,xie2026position,zheng2026group} focus optimization on student rollout prefixes.
In parallel, on-policy self-distillation (OPSD)~\cite{lu2026self,yang2026self,pan2026rlcsd,yuan2026vision} uses the same model as the teacher, conditioned on privileged context such as ground-truth answers.
%
%
TOPD~\cite{ren2026trace} extends OPD to BDLMs~\cite{cheng2026sdar} with small block sizes.
We investigate instability at larger student block sizes and address context misalignment and biased optimization dynamics.

\textbf{Efficient RLVR via Rollout Replay.}
%
To improve data efficiency in RLVR, recent works incorporate rollout replay into GRPO-style optimization, curbing policy drift via on/off-policy interleaving~\cite{zhang2025rlep,zhan2026exgrpo,li2025repo}, quality filtering~\cite{zhang2025rlep,wang2025eframe}, or importance-sampling corrections~\cite{mao2026rlvr}. 
However, these methods require delicate heuristics to manage staleness and off-policy bias. In contrast, we extend rollout replay to OPD, which exhibits fundamentally higher tolerance to trajectory staleness. 
This allows \mbox{Un-OPD} to employ an aggressive replay schedule that substantially reduces rollout generation overhead without the complexity of trace filtering or data mixing.

\section{Method}
\label{sec:method}

We propose \mbox{\textbf{Un-OPD}}, a framework that scales OPD to BDLMs with large block sizes by integrating two recipes: context-aligned distillation and support-rebalanced distillation.
%
Our design is driven by two key sources of training instability: \textit{context misalignment} arising from mismatched block boundaries and \textit{biased learning dynamics} inherent to standard OPD.
In this section, we first provide the preliminaries in Sec.~\ref{sec:Preliminaries}. Next, Sec.~\ref{sec:context_aligned_distillation} analyzes context misalignment and introduces context-aligned distillation.
Then, Sec.~\ref{sec:agreement_calibrated_distillation} investigates the biased learning dynamics of OPD and introduces our support rebalance strategy. Finally, Sec.~\ref{sec:overall_objective} unifies these components into the overall Un-OPD training framework while introducing rollout reuse to reduce trajectory-sampling overhead.

\subsection{Preliminaries}
\label{sec:Preliminaries}
\textbf{On-policy distillation.}
On-policy distillation (OPD) trains a student policy $\pi_{\theta}$ on its own rollouts using token-level supervision from a fixed teacher policy $\pi_{\mathrm{T}}$. Given a prompt $q\sim\mathcal{D}$, the student generates a response $\tau=(o_1,\ldots,o_{|\tau|})$ where $|\tau|$ is the length of the trajectory. Let $c_i$ denote the conditioning context at which token $o_i$ is predicted, including the prompt and the visible decoding state. In the reverse-KL formulation considered here, the student minimizes:
\begin{equation}
    \mathcal{L}_{\mathrm{OPD}}(\theta)
    = \mathbb{E}_{q\sim\mathcal{D},\,\tau\sim\pi_{\theta}(\cdot\mid q)}
    \left[\frac{1}{|\tau|}\sum_{i=1}^{|\tau|}
    D_{\mathrm{KL}}\!\left(\pi_{\theta}(\cdot\mid c_i)\,\Vert\,\pi_{\mathrm{T}}(\cdot\mid c_i)\right)\right],
    \label{eq:opd_objective}
\end{equation}
where sampled rollout contexts are held fixed during each optimization update. The reverse KL at position $i$ is computed over the vocabulary $\mathcal{V}$:
\begin{equation}
    \mathcal{L}_i
    = \sum_{v\in\mathcal{V}}\pi_{\theta}(v\mid c_i)
    \bigl[\log\pi_{\theta}(v\mid c_i)-\log\pi_{\mathrm{T}}(v\mid c_i)\bigr].
    \label{eq:opd_token_rkl}
\end{equation}
%
%

%
\textbf{Trace-aligned OPD for block diffusion language models (BDLMs).}
BDLMs generate sequences in a semi-autoregressive fashion: they proceed block by block, iteratively revealing tokens within the current block while conditioning on all preceding, fully generated blocks.
%
\begin{figure}[t]
    \centering
    \edef\contextfigureheight{\the\dimexpr0.25\textwidth\relax}
    \begin{subfigure}[t]{0.52\textwidth}
        \vspace{0pt}
        \centering
        \includegraphics[height=\contextfigureheight]{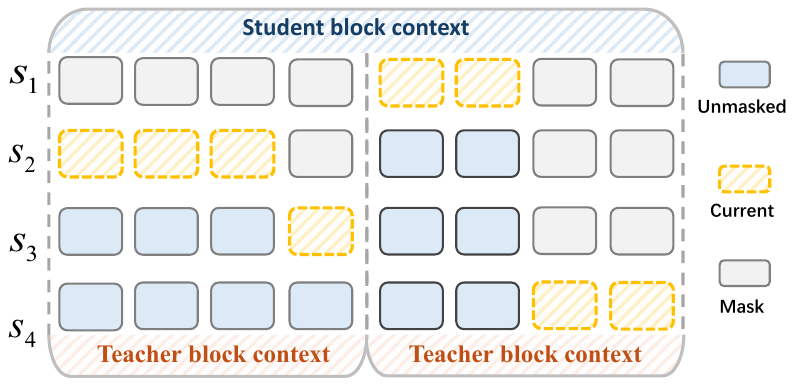}
        \caption{}
        \label{fig:Teacher_Student_Context_Window}
    \end{subfigure}
    \hfill
    \begin{subfigure}[t]{0.46\textwidth}
        \vspace{0pt}
        \centering
        \includegraphics[height=\contextfigureheight]{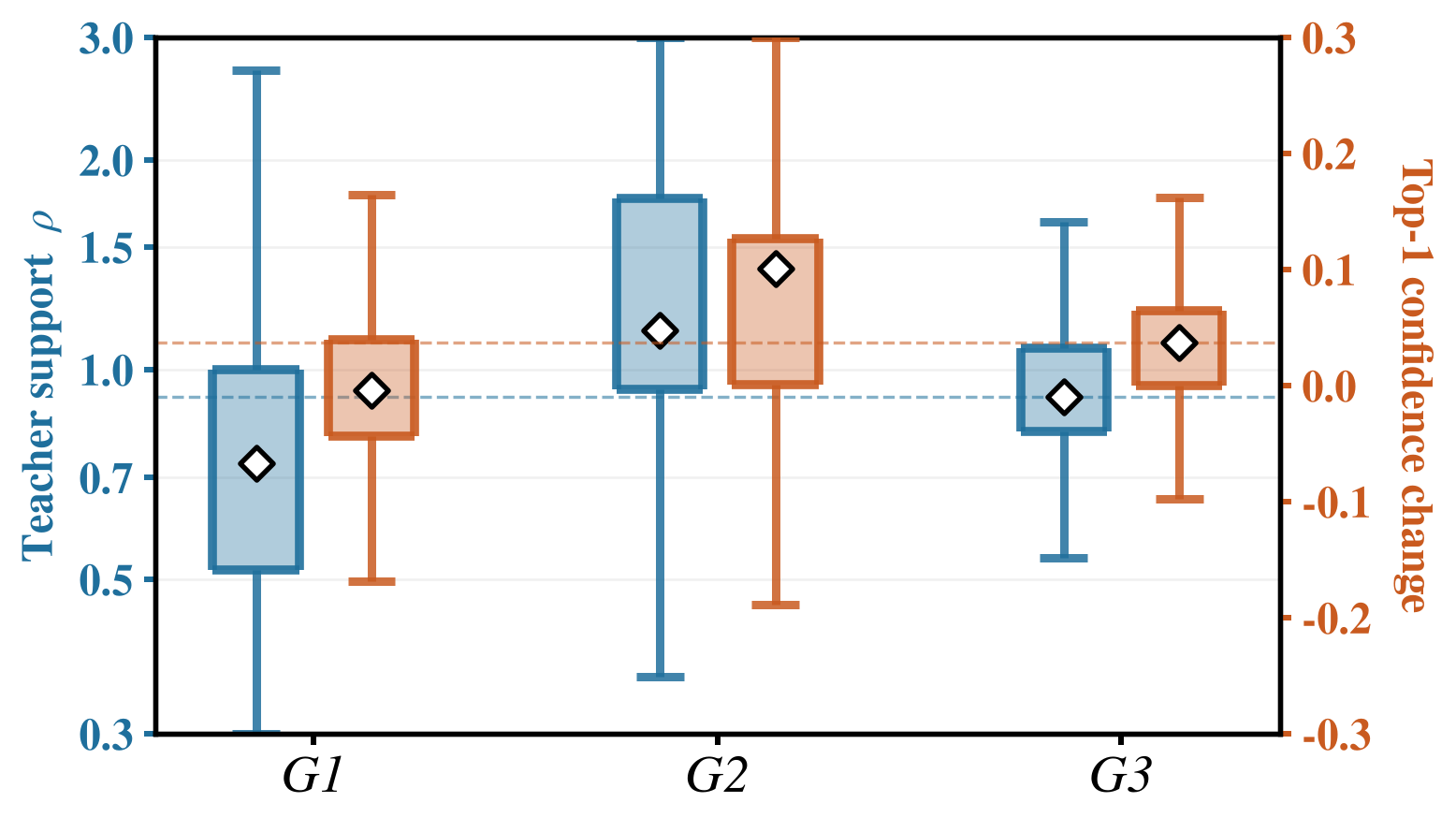}
        \caption{}
        \label{fig:G1_G2_G3}
    \end{subfigure}
    \vspace{-3mm}
    \caption{
        \textbf{Context misalignment and its effects on OPD.}
        \textbf{(a)} The student denoises an 8-token block, while the teacher provides supervision within 4-token blocks. Contexts are misaligned at states $s_1$--$s_3$ and aligned at $s_4$.
        \textbf{(b)} Box plots show distributions of teacher support $\rho$ (blue) and post-OPD top-1 probability change (orange) for context regimes $G_1$, $G_2$, and $G_3$. The diamonds mark mean values.
    }
    \label{fig:context_misalignment_analysis}
\end{figure}
Following TOPD~\cite{ren2026trace} and TraceRL~\cite{wang2026revolutionizing} (a GRPO-based method for block diffusion language models), we tailor OPD to this decoding paradigm by supervising the student directly along its on-policy denoising trajectory.
Specifically, given a prompt $q$, the student samples an execution trace $\xi=(s_0, a_0, s_1, \ldots, a_{\mathrm{K}-1}, s_{\mathrm{K}})$, where $s_k$ denotes the intermediate decoding state (with partially unmasked tokens), $a_k=\{(i, x_i)\}$ represents the set of token decisions committed at step $k$, and $s_{\mathrm{K}}$ is the finalized sequence.
For each committed position $(i,x_i)\in a_k$, we evaluate
$\pi_{\mathrm{T}}(\cdot\mid q,s_k,i)$ and
$\pi_\theta(\cdot\mid q,s_k,i)$ and minimize their reverse-KL divergence.

\subsection{Context Aligned Distillation}
\label{sec:context_aligned_distillation}

\textbf{Understanding context misalignment.}
Unlike autoregressive teacher--student pairs with identical causal contexts,
BDLMs with different block sizes can have misaligned visible contexts even under trace-aligned OPD.
Fig.~\ref{fig:context_misalignment_analysis}(\subref{fig:Teacher_Student_Context_Window}) illustrates this mismatch for teacher and student block sizes of $4$ and $8$.
The student undergoes four commit steps within its $8$-token block (yielding states $s_1$--$s_4$), whereas the teacher's span is split into two $4$-token blocks by the red dashed boundary. Consequently, their contexts align only at $s_4$, while $s_1$--$s_3$ suffer from distinct mismatches: at $s_1$, the teacher conditions on the fully generated left-block context that remains noisy to the student, whereas at $s_2$ and $s_3$, the student accesses unmasked context from the right-block that lies strictly beyond the teacher's first block boundary.

\textbf{Optimization bias under context misalignment.}
To probe how context misalignment affects student optimization, we pair an $\texttt{SDAR-4B-b8}$ student with a $\texttt{TraDo-8B-b4}$ teacher and categorize decoding states into three regimes: $\mathit{G}_1$ (student sees more context such as $s_2, s_3$), $\mathit{G}_2$ (teacher sees more such as $s_1$), and $\mathit{G}_3$ (contexts are aligned such as $s_4$).
Let $h_i=\arg\max_v p_{\mathrm{S},i}(v)$ denote the student's top-$1$ token at position $i$. We quantify the teacher support $\rho_i=p_{\mathrm{T},i}(h_i)/p_{\mathrm{S},i}(h_i)$, where $\rho_i>1$ and $\rho_i<1$ denote stronger and weaker teacher support for the student's top-$1$ prediction, respectively.
In Fig.~\ref{fig:context_misalignment_analysis}(\subref{fig:G1_G2_G3}), the blue box plots show the teacher support distributions across the three state groups, while the orange box plots show the distributions of changes in the student's top-1 probability after training with OPD exclusively on each group. The teacher support skews predominantly below $1.0$ in $\mathit{G}_1$ and above $1.0$ in $\mathit{G}_2$, whereas $\mathit{G}_3$ remains well-balanced.
The confidence change mirrors this pattern: $\mathit{G}_1$ yields negligible confidence change, $\mathit{G}_2$ causes a sharp surge, and $G_3$ produces a moderate rise.
These results demonstrate that context asymmetry systematically biases teacher supervision signals on the student rollout tokens in two opposite directions:
\begin{itemize}[leftmargin=*, itemsep=1pt, topsep=1pt]
    \item \textbf{Under-supportive supervision signals.} With less visible context than the student, the teacher tends to underestimate the student's valid rollouts relative to aligned contexts, which leaves the student at risk of \textbf{failing to reinforce confidence in correct predictions} through OPD.
    \item \textbf{Over-supportive supervision signals.} With more visible context than the student, the teacher tends to provide overly aggressive support relative to aligned contexts, which puts the student at risk of \textbf{becoming overconfident in incorrect predictions} through OPD.
\end{itemize}

\textbf{Context aligned step selection.}
%
%
%
Let $s_{\mathrm{T},k}$ and $s_{\mathrm{S},k}$ denote the teacher and student block states at step $k$.
To mitigate context-induced bias, we retain only decisions committed at steps with aligned visible contexts:
\begin{equation}
    \mathcal{A}_{\mathrm{align}}
    =\bigl\{(k,i):(i,x_i)\in a_k,\ 0\le k<K,\ s_{\mathrm{T},k}=s_{\mathrm{S},k}\bigr\}.
    \label{eq:aligned_decision_set}
\end{equation}
Here, $(k,i)$ denotes position $i$ committed at step $k$. Averaging the reverse KL over these positions gives the context-aligned OPD objective:
\begin{equation}
    \mathcal{L}_{\mathrm{aligned\text{-}OPD}}
    =\frac{1}{|\mathcal{A}_{\mathrm{align}}|}
    \sum_{(k,i)\in\mathcal{A}_{\mathrm{align}}}
    D_{\mathrm{KL}}\!\left(
        \pi_{\theta}(\cdot\mid q,s_{\mathrm{S},k},i)
        \,\middle\Vert\,
        \pi_{\mathrm{T}}(\cdot\mid q,s_{\mathrm{T},k},i)
    \right).
    \label{eq:aligned_opd_objective}
\end{equation}

\subsection{Support-Rebalanced Distillation}
\label{sec:agreement_calibrated_distillation}
%
%
Though we have identified and mitigated context-induced bias in Sec.~\ref{sec:context_aligned_distillation}, training remains suboptimal. Crucially, we observe that another bias, the internal optimization dynamics of OPD, also compromises BDLM training stability.
Here, we first present empirical observations of this behavior in OPD, then provide a theoretical grounding, and finally introduce a simple yet effective solution.
%

\begin{wrapfigure}{r}{0.53\textwidth}
    \vspace{-6mm}
    \centering
    \includegraphics[width=\linewidth]{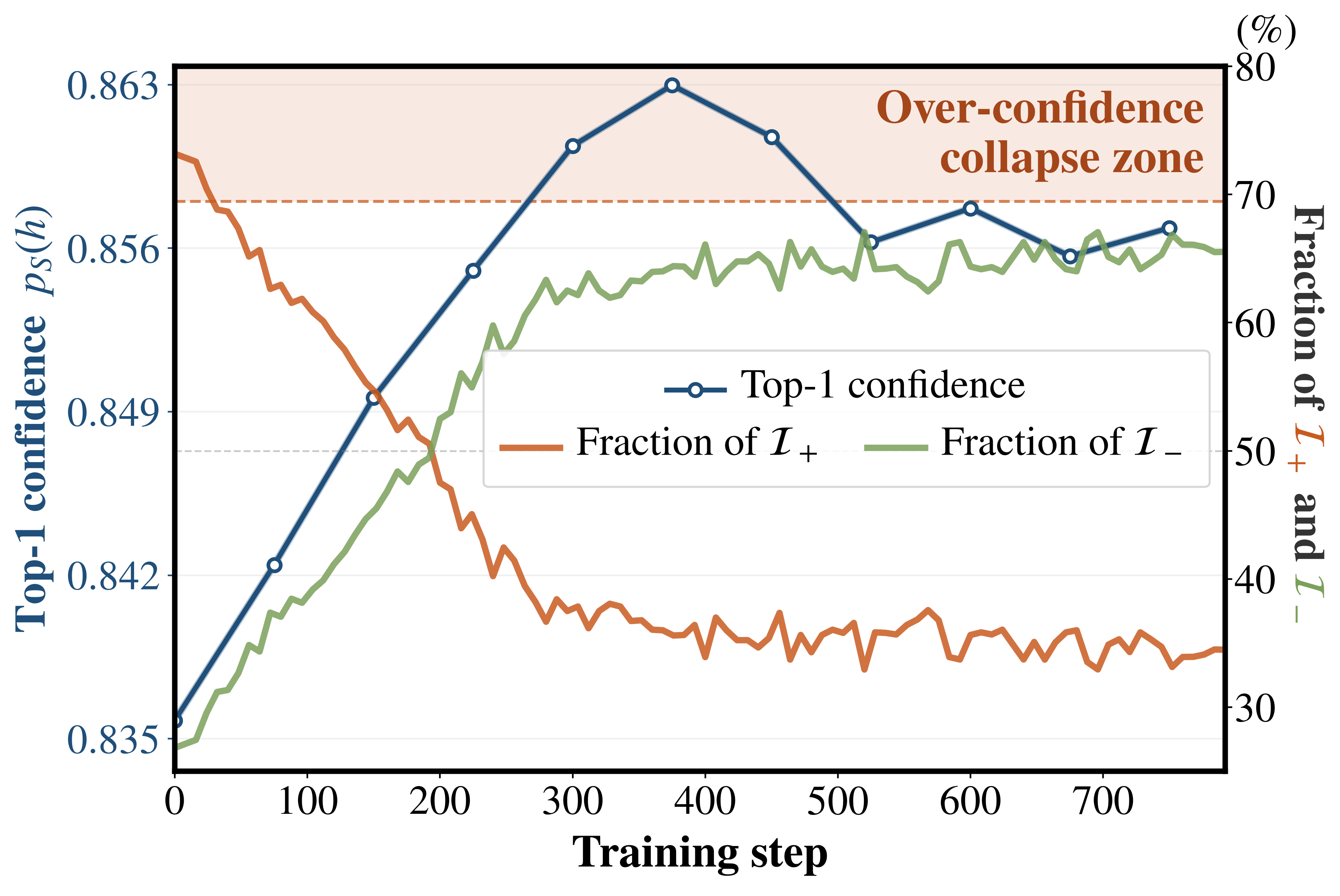}
    \vspace{-6mm}
    \caption{
        Evolution of student mean top-1 confidence (blue) alongside subset fractions of $\mathcal{I}_{+}$ (orange) and $\mathcal{I}_{-}$ (green). The shaded area denotes the overconfidence collapse zone.
    }
    \label{fig:agreement_ratio_trend}
    \vspace{-4mm}
\end{wrapfigure}

\textbf{Biased optimization dynamics in OPD.}
Since top-1 token confidence critically dictates decoding decisions in BDLMs, we begin by examining the evolution of the student's average top-1 confidence throughout the entire OPD training process.
As illustrated in Fig.~\ref{fig:agreement_ratio_trend}, decoding confidence of the student model exhibits a
distinct two-stage pattern throughout training: \textit{a rapid surge during the early phase, followed by a slight decline during later training.}
To dissect the underlying mechanism behind this phenomenon, we leverage the previously defined teacher support $\rho_i$ in Sec.~\ref{sec:context_aligned_distillation} given its tight correlation with confidence shifts, and divide the supervised positions into two distinct groups: $\mathcal{I}_{+}=\{i:\rho_i>1\}$ and $\mathcal{I}_{-}=\{i:\rho_i<1\}$.
Empirically, updates from $\mathcal{I}_{+}$ tend to increase the student\textquotesingle s decoding confidence, whereas updates from $\mathcal{I}_{-}$ tend to decrease it.
%

%
As shown in Fig.~\ref{fig:agreement_ratio_trend}, we reveal that the fraction of high-support positions steadily decreases, while that of low-support positions rises before eventually plateauing.
We hypothesize that this early confidence surge stems from an asymmetric learning dynamic: the student \textbf{rapidly internalizes} supervisory signals from high-support positions ($\mathcal{I}_{+}$), whereas updates from low-support positions ($\mathcal{I}_{-}$) \textbf{suffer from an optimization lag}. We term this learning asymmetry the \textit{intrinsic optimization bias} of OPD.
Consequently, as the student's confidence is prematurely driven up, the rollout policy becomes increasingly prone to sampling tokens beyond the teacher's support region, which in turn produces more low-support positions \ryn{that} ultimately dominate training and trigger a subsequent pullback in confidence.
\textbf{Theoretical grounding.}
To theoretically characterize this optimization bias, we examine the reverse-KL gradient update to the logit of the student's top-1 prediction $h_i$. Let $a_i=p_{\mathrm{S},i}(h_i)$ and $b_i=p_{\mathrm{T},i}(h_i)$, such that $\rho_i=b_i/a_i$. We denote by $D_i^{\mathrm{tail}}\ge0$ the reverse-KL divergence between the renormalized tail distributions of the student and teacher over all non-top-1 tokens ($v \neq h_i$).
\begin{proposition}[Asymmetric top-1 logit updates under reverse-KL]
\label{prop:agreement_head_gradient}
For a fixed context and a teacher distribution with positive token probabilities, a reverse-KL gradient descent step with learning rate $\eta>0$ directly on the student's softmax logits yields:
\begin{equation}
    \Delta z_{\mathrm{S},i}(h_i)
    =\eta a_i(1-a_i)\left[\log\frac{b_i(1-a_i)}{a_i(1-b_i)}+D_i^{\mathrm{tail}}\right].
    \label{eq:agreement_head_update}
\end{equation}
The proof is provided in Appendix~\ref{app:proof_top1_logit_asymmetry}.
\end{proposition}
Proposition~\ref{prop:agreement_head_gradient} shows that the nonnegative tail term $D_i^{\mathrm{tail}}$ reinforces upward logit updates for $\rho_i>1$, but can weaken or reverse downward updates for $\rho_i<1$.
This local logit-level analysis provides a mechanism consistent with the observed asymmetry, although actual training dynamics also depend on the overall training data distribution.

\textbf{Support-rebalanced confidence calibration.}
Due to this optimization bias, OPD is not a free lunch for BDLMs. When weaker students are guided by a strong teacher, a premature surge in confidence readily pushes them into a degenerate generation regime (Fig.~\ref{fig:agreement_ratio_trend}). Consequently, such students slip into a catastrophic overconfidence zone, failing to recover even in later training stages to a confidence regime commensurate with their actual capacity.
Here, we propose calibrating the supervision signals specifically at high-support positions.
In implementation, we temper the teacher target by interpolating its logits with the student's detached logits for each $i \in \mathcal{I}_{+}$, thereby moderating update intensity at high-support positions.
Let $\mathcal{I}$ denote the set of supervised token positions and $\lambda \in [0, 1]$ be the teacher interpolation weight at high-support positions. The calibrated teacher target and the resulting reverse-KL objective are defined as:
\begin{equation}
    \begin{gathered}
        \widetilde{z}_{\mathrm{T},i}(v)
        = \bigl[1-(1-\lambda)\mathbf{1}_{\{i\in\mathcal{I}_{+}\}}\bigr]z_{\mathrm{T},i}(v)
        +(1-\lambda)\mathbf{1}_{\{i\in\mathcal{I}_{+}\}}\operatorname{sg}\!\left[z_{\mathrm{S},i}(v)\right],\\
        \mathcal{L}_{\lambda}
        =\frac{1}{|\mathcal{I}|}\sum_{i\in\mathcal{I}}
        D_{\mathrm{KL}}\!\left(p_{\mathrm{S},i}\,\middle\Vert\,
        \operatorname{softmax}(\widetilde{z}_{\mathrm{T},i})\right).
    \end{gathered}
    \label{eq:agreement_reweighted_objective}
\end{equation}
Here, $\mathbf{1}_{\{i\in\mathcal{I}_{+}\}}$ indicates a high-support position and $\operatorname{sg}[\cdot]$ denotes stop-gradient. At these positions, the teacher and detached student logits receive weights $\lambda$ and $1-\lambda$, respectively; smaller $\lambda$ yields stronger calibration. Low-support positions retain the original teacher target.
%

\subsection{Overall Framework}
\label{sec:overall_objective}

\textbf{Overall objective.}
Unifying the two training recipes, we can obtain the unbiased OPD framework for BDLMs, whose overall objective minimizes the reverse-KL divergence over aligned decisions in $\mathcal{A}_{\mathrm{align}}$ using the calibrated teacher target:
\begin{equation}
    \mathcal{L}_{\mathrm{Un\text{-}OPD}}(\theta)
    =\mathbb{E}_{q,\xi}\!\left[
        \frac{1}{|\mathcal{A}_{\mathrm{align}}|}
        \sum_{(k,i)\in\mathcal{A}_{\mathrm{align}}}
        D_{\mathrm{KL}}\!\left(
            p_{\mathrm{S},k,i}
            \,\middle\Vert\,
            \operatorname{softmax}(\widetilde{z}_{\mathrm{T},k,i})
        \right)
    \right],
    \label{eq:un_opd_objective}
\end{equation}
where $p_{\mathrm{S},k,i}=\pi_\theta(\cdot\mid q,s_{\mathrm{S},k},i)$ and $\widetilde{z}_{\mathrm{T},k,i}$ is the calibrated target from Eq.~\ref{eq:agreement_reweighted_objective}.
When $\lambda = 1$, Eq.~\ref{eq:un_opd_objective} recovers context-aligned OPD (Eq.~\ref{eq:aligned_opd_objective}), while $\lambda < 1$ calibrates the teacher target at high-support positions to curb premature confidence inflation.

%


\textbf{Rollout reuse.}
We notice that OPD is inherently better suited for trajectory reuse compared to RLVR.
In OPD, historical rollouts serve merely as conditioning states rather than frozen discrete actions or stale scalar rewards as in RLVR.
Across multiple optimization steps on the same rollout batch, we recompute the student distributions under the current policy, while the frozen teacher distributions over the fixed states remain unchanged. The calibrated targets and distillation gradients are recomputed accordingly.
Consequently, the primary discrepancy in OPD replay arises solely from \textit{state distribution mismatch} between the sampling policy and the updated student. 
Unlike RLVR, which suffers from volatile importance sampling and reward clipping over historical rollouts, OPD suffers from substantially milder off-policy degradation.
%
This insight motivates our \textit{rollout reuse} mechanism, which performs multiple updates per rollout to effectively reduce the heavy cost of rollout sampling.
The complete Un-OPD training procedure is presented in Alg.~\ref{alg:un_opd} of appendix~\ref{app:algorithm_details}.

\section{Experiments}
\setcounter{topnumber}{2}
\renewcommand{\topfraction}{0.95}
\renewcommand{\textfraction}{0.05}
\renewcommand{\floatpagefraction}{0.8}

In this section, we conduct comprehensive experiments across BDLMs with varying block sizes
to evaluate the effectiveness of Un-OPD. 
We first detail the experimental configurations in Sec.~\ref{exp_setting}, then assess how the
proposed Un-OPD improves performance across different benchmarks in Sec.~\ref{main_exp}.
Finally, we provide an ablation study about the core recipes of Un-OPD in Sec.~\ref{ablation}.

\begingroup
\newcommand{\posdelta}[1]{\ensuremath{_{\color{green!60!black}\scriptscriptstyle +#1}}}
\newcommand{\negdelta}[1]{\ensuremath{_{\color{red!80!black}\scriptscriptstyle -#1}}}
\newcommand{\zerodelta}{\ensuremath{_{\color{black!55}\scriptscriptstyle +0.00}}}

\begin{table*}[t]
  \centering
  \caption{
    Performance comparison on math and code generation
  benchmarks. We report accuracy (\%) with gains over the base model computed by direct subtraction.
With the teacher block size fixed at 4, student block sizes of 8 and 4 correspond to \textbf{asymmetric} and \textbf{symmetric} configurations, respectively.
  The best results in each group are \textbf{boldfaced} and \underline{underlined}.
  }
  \label{tab:math_and_code}
  \setlength{\tabcolsep}{4.5pt}
  \renewcommand{\arraystretch}{1.08}
  \resizebox{0.95\textwidth}{!}{%
  \begin{tabular}{@{}l|ccccc|ccc@{}}
    \toprule
    \textbf{Method}
      & \multicolumn{5}{c|}{\textbf{Mathematical Reasoning}}
      & \multicolumn{3}{c}{\textbf{Code Generation}} \\
    \cmidrule(lr){2-6}\cmidrule(l){7-9}
      & \textbf{GSM8K} & \textbf{MATH500} & \textbf{AMC2023} & \textbf{Olympiad} & \textbf{Avg.}
      & \textbf{MBPP} & \textbf{HumanEval} & \textbf{Avg.} \\
    \midrule
    \multicolumn{9}{c}{\textit{Student Model: SDAR-4B-b8}} \\
    \midrule
    \rowcolor{gray!15}
    Base
      & 88.9 
      & 66.3 
      & 42.5 
      & 29.2 
      & 56.7 
      & 53.5 
      & 67.9 
      & 60.7 \\
    + SFT
      & $88.1\negdelta{0.8}$
      & $65.4\negdelta{0.9}$
      & $40.3\negdelta{2.2}$
      & $27.9\negdelta{1.3}$
      & $55.4\negdelta{1.3}$
      & $51.7\negdelta{1.8}$
      & $65.5\negdelta{2.4}$
      & $58.6\negdelta{2.1}$ \\
    + TraceRL
      & $89.5\posdelta{0.6}$ 
      & $69.4\posdelta{3.1}$ 
      & $\underline{\textbf{46.8}}\posdelta{4.3}$ 
      & $30.0\posdelta{0.8}$ 
      & $58.9\posdelta{2.2}$ 
      & $54.1\posdelta{0.6}$ 
      & $68.9\posdelta{1.0}$ 
      & $61.5\posdelta{0.8}$ \\
    + TOPD
      & $88.4\negdelta{0.5}$
      & $67.1\posdelta{0.8}$
      & $43.6\posdelta{1.1}$
      & $30.8\posdelta{1.6}$
      & $57.5\posdelta{0.8}$
      & $52.3\negdelta{1.2}$
      & $65.1\negdelta{2.8}$
      & $58.7\negdelta{2.0}$ 
      \\
    \rowcolor{blue!10}
    + Un-OPD (Ours)
      & $\underline{\textbf{90.0}}\posdelta{1.1}$
      & $\underline{\textbf{70.1}}\posdelta{3.8}$
      & $45.8\posdelta{3.3}$
      & $\underline{\textbf{32.8}}\posdelta{3.6}$
      & $\underline{\textbf{59.7}}\posdelta{3.0}$
      & $\underline{\textbf{55.6}}\posdelta{2.1}$
      & $\underline{\textbf{71.1}}\posdelta{3.2}$
      & $\underline{\mathbf{63.4}}\posdelta{2.7}$ \\
    \midrule
    \multicolumn{9}{c}{\textit{Student Model: SDAR-1.7B-b8}} \\
    \midrule
    \rowcolor{gray!15}
    Base
      & 74.8 
      & 51.9 
      & 25.9 
      & 17.3 
      & 42.5 
      & 41.2 
      & 52.4 
      & 46.8 \\
    + SFT
      & $73.4\negdelta{1.4}$ 
      & $50.1\negdelta{1.8}$ 
      & $23.3\negdelta{2.6}$ 
      & $15.7\negdelta{1.6}$ 
      & $40.6\negdelta{1.9}$ 
      & $40.5\negdelta{0.7}$ 
      & $53.5\posdelta{1.1}$ 
      & $47.0\posdelta{0.2}$ 
      \\
    + TraceRL
      & $\underline{\textbf{75.6}}\posdelta{0.8}$ 
      & $53.0\posdelta{1.1}$ 
      & $\underline{\textbf{27.8}}\posdelta{1.9}$ 
      & $18.3\posdelta{1.0}$ 
      & $43.7\posdelta{1.2}$ 
      & $\underline{\textbf{43.5}}\posdelta{2.3}$ 
      & $54.4\posdelta{2.0}$
      & $\underline{\textbf{49.0}}\posdelta{2.2}$ \\
    + TOPD
      & $74.4\negdelta{0.4}$ 
      & $50.9\negdelta{1.0}$ 
      & $25.5\negdelta{0.4}$ 
      & $17.2\negdelta{0.1}$ 
      & $42.0\negdelta{0.5}$ 
      & $40.1\negdelta{1.1}$ 
      & $50.9\negdelta{1.5}$ 
      & $45.5\negdelta{1.3}$ \\
    \rowcolor{blue!10}
    + Un-OPD (Ours)
      & $75.3\posdelta{0.5}$ 
      & $\underline{\textbf{54.9}}\posdelta{3.0}$ 
      & $27.7\posdelta{1.8}$ 
      & $\underline{\textbf{18.6}}\posdelta{1.3}$ 
      & $\underline{\textbf{44.1}}\posdelta{1.6}$ 
      & $42.4\posdelta{1.2}$ 
      & $\underline{\textbf{55.6}}\posdelta{3.2}$ 
      & $\underline{\mathbf{49.0}}\posdelta{2.2}$
      \\
    \midrule
    \multicolumn{9}{c}{\textit{Student Model: SDAR-4B-b4}} \\
    \midrule
    \rowcolor{gray!15}
    Base
      & 88.1 
      & 66.5 
      & 43.1 
      & 29.5 
      & 56.8 
      & 50.8 
      & 65.2
      & 58.0 \\
    + SFT
      & $\underline{\textbf{89.8}}\posdelta{1.7}$
      & $67.4\posdelta{0.9}$
      & $40.2\negdelta{2.9}$
      & $31.7\posdelta{2.2}$
      & $57.3\posdelta{0.5}$
      & $54.8\posdelta{4.0}$
      & $\underline{\textbf{71.3}}\posdelta{6.1}$
      & $63.1\posdelta{5.1}$ \\
    + TraceRL
      & $87.1\negdelta{1.0}$
      & $70.4\posdelta{3.9}$
      & $44.2\posdelta{1.1}$
      & $30.7\posdelta{1.2}$
      & $58.1\posdelta{1.3}$
      & $56.4\posdelta{5.6}$
      & $71.2\posdelta{6.0}$
      & $63.8\posdelta{5.8}$ \\
    + TOPD
      & $89.7\posdelta{1.6}$
      & $70.8\posdelta{4.3}$
      & $45.0\posdelta{1.9}$
      & $32.9\posdelta{3.4}$
      & $59.6\posdelta{2.8}$
      & $56.8\posdelta{6.0}$
      & $70.8\posdelta{5.6}$
      & $63.8\posdelta{5.8}$
      \\
    \rowcolor{blue!10}
    + Un-OPD (Ours)
      & $89.5\posdelta{1.4}$
      & $\underline{\textbf{72.0}}\posdelta{5.5}$
      & $\underline{\textbf{45.4}}\posdelta{2.3}$
      & $\underline{\textbf{33.5}}\posdelta{4.0}$
      & $\underline{\textbf{60.1}}\posdelta{3.3}$
      & $\underline{\textbf{57.1}}\posdelta{6.3}$
      & $71.0\posdelta{5.8}$
      & $\underline{\textbf{64.1}}\posdelta{6.1}$ \\
    \midrule
    \multicolumn{9}{c}{\textit{Student Model: SDAR-1.7B-b4}} \\
    \midrule
    \rowcolor{gray!15}
    Base
      & 74.4 
      & 53.2 
      & 26.7 
      & 16.8 
      & 42.8 
      & 41.2 
      & 53.0 
      & 47.1 
      \\
    + SFT
      & $76.4\posdelta{2.0}$ 
      & $54.2\posdelta{1.0}$ 
      & $28.8\posdelta{2.1}$ 
      & $19.5\posdelta{2.7}$ 
      & $44.7\posdelta{1.9}$ 
      & $43.9\posdelta{2.7}$ 
      & $56.1\posdelta{3.1}$ 
      & $50.0\posdelta{2.9}$ 
      \\
    + TraceRL
      & $\underline{\textbf{76.7}}\posdelta{2.3}$ 
      & $55.9\posdelta{2.7}$ 
      & $30.6\posdelta{3.9}$ 
      & $19.0\posdelta{2.2}$ 
      & $45.6\posdelta{2.8}$ 
      & $45.4\posdelta{4.2}$ 
      & $55.9\posdelta{2.9}$ 
      & $50.7\posdelta{3.6}$ 
      \\
    + TOPD
      & $76.6\posdelta{2.2}$ 
      & $57.7\posdelta{4.5}$ 
      & $30.6\posdelta{3.9}$ 
      & $21.1\posdelta{4.3}$ 
      & $46.5\posdelta{3.7}$ 
      & $43.8\posdelta{2.6}$ 
      & $57.4\posdelta{4.4}$ 
      & $50.6\posdelta{3.5}$ 
      \\
    \rowcolor{blue!10}
    + Un-OPD (Ours)
      & $76.1\posdelta{1.7}$ 
      & $\underline{\textbf{58.8}}\posdelta{5.6}$ 
      & $\underline{\textbf{31.5}}\posdelta{4.8}$ 
      & $\underline{\textbf{21.6}}\posdelta{4.8}$ 
      & $\underline{\textbf{47.0}}\posdelta{4.2}$ 
      & $\underline{\textbf{46.1}}\posdelta{4.9}$ 
      & $\underline{\textbf{57.9}}\posdelta{4.9}$ 
      & $\underline{\textbf{52.0}}\posdelta{4.9}$ 
      \\
    \bottomrule
  \end{tabular}%
  }
\end{table*}
\endgroup

%
\newcommand{\tabposdelta}[1]{\textcolor{green!60!black}{\scriptstyle +#1}}
\newcommand{\tabnegdelta}[1]{\textcolor{red!80!black}{\scriptstyle -#1}}
\begin{table*}[t]
    \centering
    \scriptsize
    \renewcommand{\arraystretch}{1.1}
    \setlength{\tabcolsep}{2.8pt}
    \caption{Ablation of Un-OPD components on SDAR-4B-b8 and SDAR-1.7B-b8.
    We report accuracy (\%) on three mathematical reasoning benchmarks and the corresponding training cost in GPU hours.
    Components are added cumulatively; subscripts show accuracy changes from the preceding row, computed by direct subtraction.}
    \label{tab:suffix_strategy}
    \resizebox{0.98\textwidth}{!}{%
    \begin{tabular}{@{}l|cccc|cccc@{}}
        \toprule
        \multirow{2}{*}{\textbf{Method}}
        & \multicolumn{4}{c|}{\textbf{SDAR-4B-b8}}
        & \multicolumn{4}{c}{\textbf{SDAR-1.7B-b8}} \\
        \cmidrule(lr){2-5}\cmidrule(l){6-9}
        & \textbf{MATH500} & \textbf{AMC2023} & \textbf{Olympiad} & \textbf{GPU hours}
        & \textbf{MATH500} & \textbf{AMC2023} & \textbf{Olympiad} & \textbf{GPU hours} \\
        \midrule
        Base                    
        & 66.3 
        & 42.5 
        & 29.2 
        & -- 
        & 51.9 
        & 25.9 
        & 17.3 
        & -- 
        \\
        + TOPD                   
        & $67.1_{\tabposdelta{0.8}}$ 
        & $43.6_{\tabposdelta{1.1}}$ 
        & $30.8_{\tabposdelta{1.6}}$ 
        & $152$ 
        & $50.9_{\tabnegdelta{1.0}}$ 
        & $25.5_{\tabnegdelta{0.4}}$ 
        & $17.2_{\tabnegdelta{0.1}}$ 
        & $104$ 
        \\
        + context-aligned       
        & $69.5_{\tabposdelta{2.4}}$ 
        & $44.5_{\tabposdelta{0.9}}$ 
        & $32.6_{\tabposdelta{1.8}}$ 
        & $152$ 
        & $51.5_{\tabposdelta{0.6}}$ 
        & $26.7_{\tabposdelta{1.2}}$ 
        & $17.3_{\tabposdelta{0.1}}$ 
        & $104$ 
        \\
        + support-rebalanced      
        & $70.3_{\tabposdelta{0.8}}$ 
        & $45.7_{\tabposdelta{1.2}}$ 
        & $33.0_{\tabposdelta{0.4}}$
        & $152$ 
        & $54.4_{\tabposdelta{2.9}}$ 
        & $27.5_{\tabposdelta{0.8}}$ 
        & $18.9_{\tabposdelta{1.6}}$ 
        & $104$ 
        \\
        + rollout reuse        
        & $70.1_{\tabnegdelta{0.2}}$ 
        & $45.8_{\tabposdelta{0.1}}$ 
        & $32.8_{\tabnegdelta{0.2}}$ 
        & $72$ 
        & $54.9_{\tabposdelta{0.5}}$ 
        & $27.7_{\tabposdelta{0.2}}$ 
        & $18.6_{\tabnegdelta{0.3}}$ 
        & $48$ 
        \\
        \bottomrule
    \end{tabular}%
    }
\end{table*}
%
%
\subsection{Experimental Settings}
\label{exp_setting}
\begin{figure*}[t]
    \centering
    \begin{subfigure}[t]{0.49\textwidth}
        \centering
        \scalebox{1}[0.96]{\includegraphics[width=\linewidth]{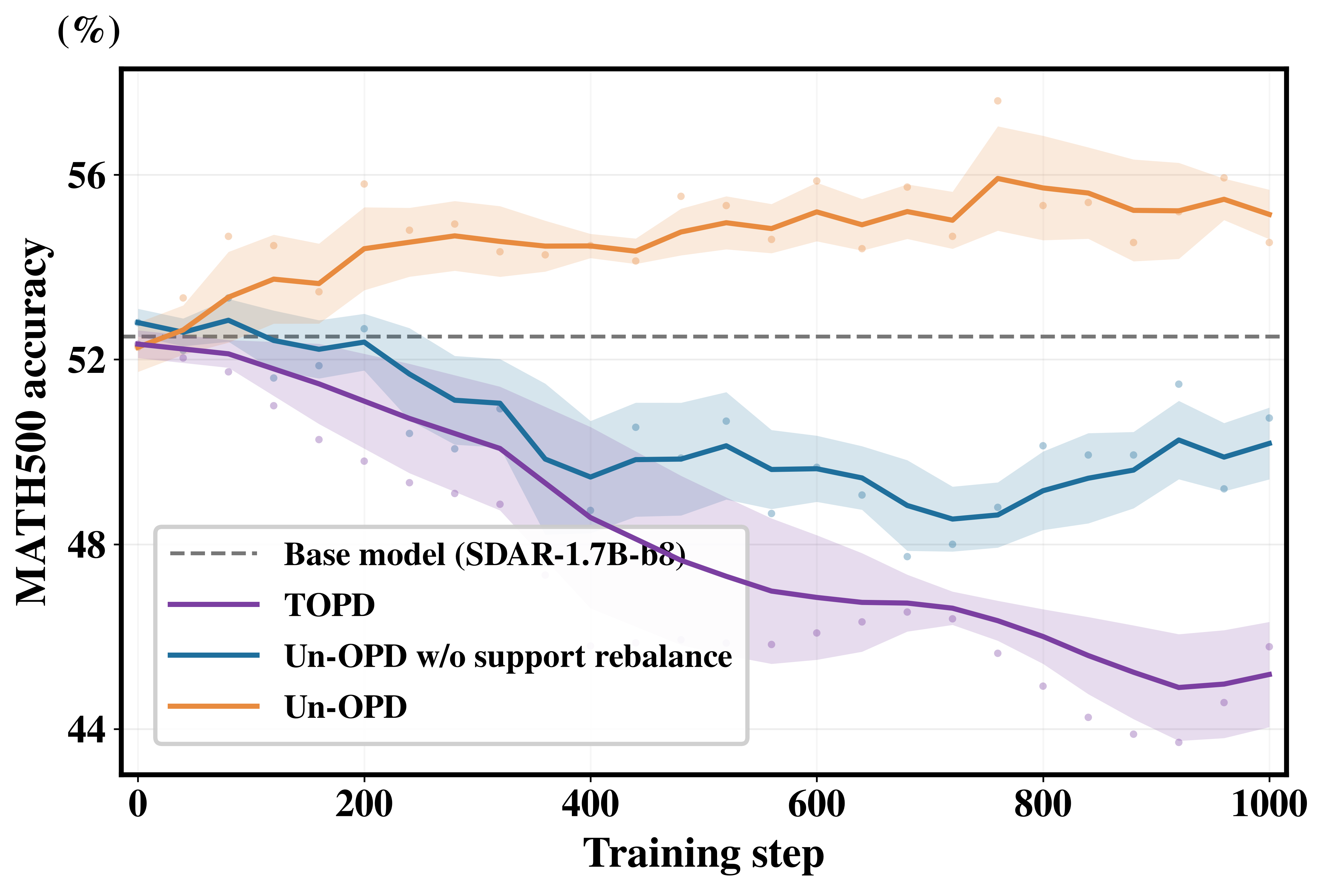}}
        \caption{SDAR-1.7B-b8.}
        \label{fig:dynamic_eval_ablation_1p7b}
    \end{subfigure}
    \hfill
    \begin{subfigure}[t]{0.49\textwidth}
        \centering
        \scalebox{1}[0.96]{\includegraphics[width=\linewidth]{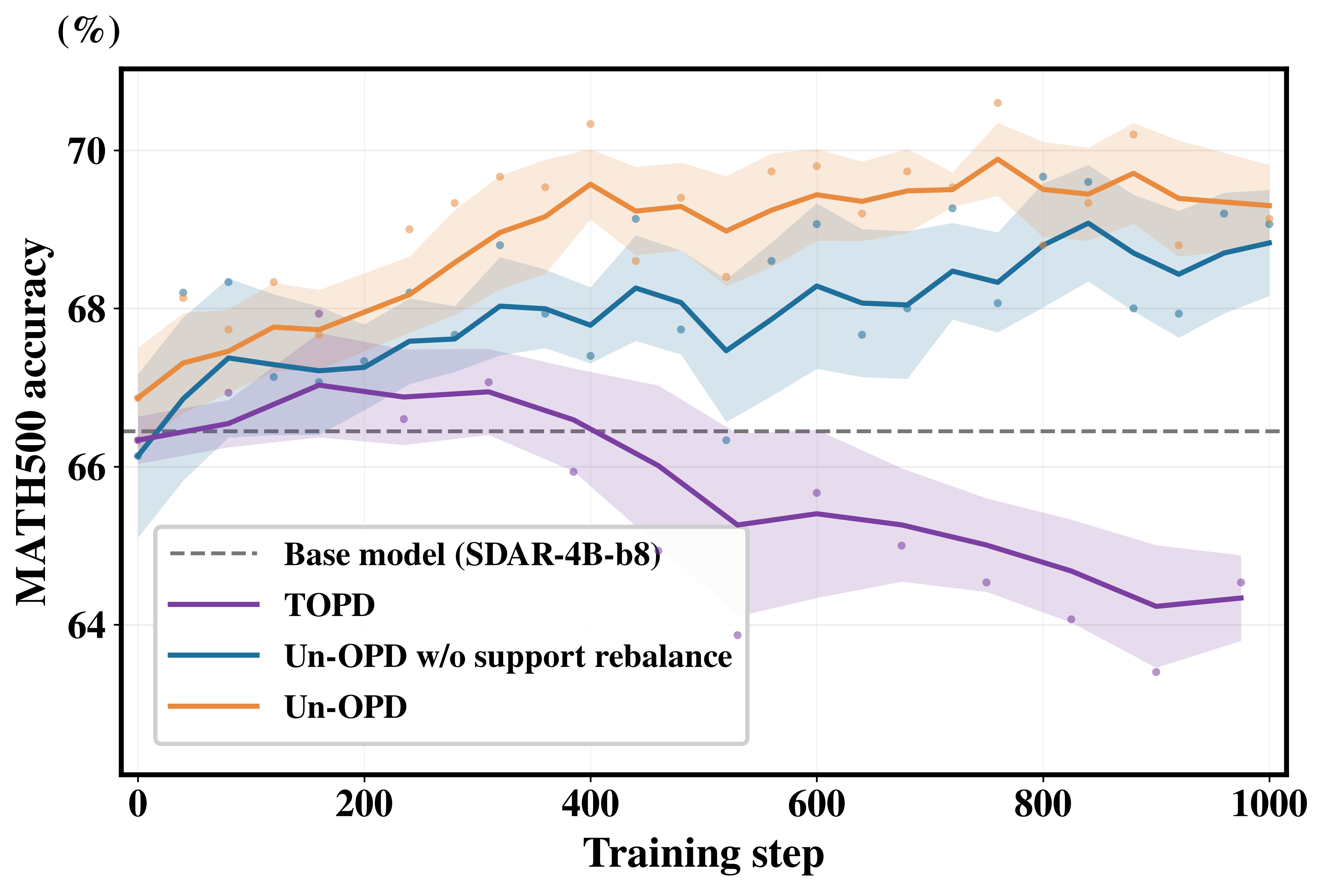}}
        \caption{SDAR-4B-b8.}
        \label{fig:dynamic_eval_ablation_4b}
    \end{subfigure}
    \caption{
        \textbf{Training dynamics on MATH500.}
        We compare the base model (dashed line), TOPD, Un-OPD without support rebalance, and the complete Un-OPD. Solid lines show the
        smoothed performance trend.
    }
    \label{fig:dynamic_eval_ablation}
    \vspace{-4mm}
\end{figure*}
\textbf{Baselines and Benchmarks.}
To comprehensively evaluate the scalability and robustness of \mbox{Un-OPD}, we employ \texttt{SDAR-4B} and \texttt{SDAR-1.7B}~\cite{cheng2026sdar} with varying block sizes ($b \in \{4, 8\}$) as student BDLMs, paired with a default \texttt{TraDo-8B}~\cite{wang2026revolutionizing} as teacher ($b=4$).
We evaluate the models on standard benchmarks for mathematical reasoning (\textit{e.g.,} GSM8K~\cite{cobbe2021training}, MATH-500~\cite{hendrycks2021measuring}, AMC 2023), as well as HumanEval~\cite{chen2021evaluating} and MBPP~\cite{austin2021program} for code generation.
To ensure fair comparisons, we evaluate against \textbf{SFT}, \textbf{TraceRL}~\cite{wang2026revolutionizing}, and \textbf{TOPD}~\cite{ren2026trace} as three primary baselines.

\textbf{Implementation details.}
For mathematical reasoning, we follow TraceRL~\cite{wang2026revolutionizing} and use 8K math problems for post-training. For code generation, we use a curated subset of over $10,000$ problems from AceCoder~\cite{zeng2025acecoder}.
Student trajectories are collected using dynamic decoding with a batch size of $64$ prompts (one rollout per prompt).
We adopt the dynamic decoding schedule with confidence threshold $0.9$ and temperature $1.0$ for evaluation unless otherwise specified.
We report mean accuracy over four sampled responses per problem ($\operatorname{avg}@4$), except for AMC 2023, where we use 16 responses per problem ($\operatorname{avg}@16$).
%
All models are trained for 1,000 gradient steps on an 8-GPU NVIDIA H20 server to guarantee convergence.
Full training, implementation, baseline, and evaluation details are provided in Appendix~\ref{app:experimental_details}.

\subsection{Main Results}
\label{main_exp}

Tab.~\ref{tab:math_and_code} shows distillation performance when scaling to students with a larger block size ($b=8$) under a teacher ($b=4$). Across both student scales, \mbox{Un-OPD} outperforms TOPD on all six benchmarks and achieves the highest or tied-highest average accuracy in both domains.
For the \texttt{SDAR-4B-b8} student, \mbox{Un-OPD} raises average accuracy in mathematical reasoning from $57.5\%$ to $59.7\%$ and in code generation from $58.7\%$ to $63.4\%$.
\begin{wraptable}{r}{0.45\textwidth}
    \centering
    \vspace{-3mm}
    \caption{Sensitivity to the interpolation coefficient $\lambda$ on MATH500.}
    \label{tab:hyperparameter_ablation}
    \small
    \setlength{\tabcolsep}{4pt}
    \renewcommand{\arraystretch}{1.1}
    \begin{tabular*}{\linewidth}{@{\extracolsep{\fill}}l|ccccc@{}}
        \toprule
        $\lambda$ & 0.2 & 0.4 & 0.6 & 0.8 & 1.0 \\
        \midrule
        SDAR-1.7B-b8 & 53.7 & 54.9 & 54.1 & 52.4 & 50.9 \\
        SDAR-4B-b8 & 70.0 & 70.1 & 69.8 & 68.7 & 67.1 \\
        \bottomrule
    \end{tabular*}
    \vspace{-2mm}
\end{wraptable}
The advantage is even more striking on the smaller \texttt{SDAR-1.7B-b8} student: while TOPD suffers from severe performance degradation and consistently falls below the base model, \mbox{Un-OPD} successfully stabilizes training, elevating math and code averages to $44.1\%$ and $49.0\%$.
Overall, these results support the effectiveness of addressing optimization biases when transferring knowledge into larger-block BDLMs.
%
%
Moreover, for BDLMs with the same block size (\textit{i.e.,} $4$), \mbox{Un-OPD} still achieves better overall performance than TOPD.
For example, the mathematical reasoning accuracy increases from $46.5\%$ to $47.0\%$ for \texttt{SDAR-1.7B-b4}, yielding a $0.5\%$ gain.
%
These results demonstrate that \mbox{Un-OPD} readily generalizes across varying block-size configurations.

\subsection{Ablation Study}
\label{ablation}

\textbf{Ablation of core components.}
Tab.~\ref{tab:suffix_strategy} ablates our proposed components on both students (\texttt{SDAR-1.7B-b8} and \texttt{SDAR-4B-b8}).
Directly applying standard OPD (TOPD) yields marginal gains on \texttt{SDAR-4B-b8} and even causes performance degradation on the smaller \texttt{SDAR-1.7B-b8}, confirming the deficiency of trace-aligned distillation.
Introducing \textbf{context-aligned distillation} improves all three reported benchmarks on both student scales, with MATH500 gains of $2.4\%$  for the  \texttt{SDAR-4B-b8}.
However, context alignment alone leaves the weaker \texttt{SDAR-1.7B-b8} student underperforming.
Adding \textbf{support-rebalanced distillation} to calibrate updates at high-support positions improves accuracy on all three reported benchmarks for the \texttt{SDAR-1.7B-b8} student (\textit{e.g.,} a gain of $2.9\%$ on MATH500).
Finally, the \textbf{rollout reuse} mechanism reduces training overhead (from $152$ to $72$ GPU hours for 4B) while largely preserving accuracy.
These ablations validate that each component is indispensable to achieving both superior accuracy and training efficiency.
%

Furthermore, Fig.~\ref{fig:dynamic_eval_ablation} compares the training dynamics on MATH500 across both student scales. While standard TOPD suffers from severe late-stage degradation, \mbox{Un-OPD} maintains steady improvements throughout training. Crucially, omitting support rebalance causes the smaller \texttt{SDAR-1.7B-b8} student to collapse below base accuracy and limits the gains of \texttt{SDAR-4B-b8}. This confirms that support-rebalanced calibration is essential to ensure robust convergence, particularly for capacity-constrained models.
%

\textbf{Hyper-parameter analysis of $\lambda$ and $N$.}
We examine the sensitivity of Un-OPD to the interpolation coefficient $\lambda$ and rollout reuse epoch number $N$ on MATH500.
For $\lambda$, Tab.~\ref{tab:hyperparameter_ablation} shows that both student scales achieve their highest reported accuracy at $\lambda=0.4$: $54.9\%$ for \texttt{SDAR-1.7B-b8} and $70.1\%$ for \texttt{SDAR-4B-b8}.
Increasing $\lambda$ beyond $0.4$ reduces accuracy on both models, supporting a moderate calibration strength.
%
\begin{figure}[t]
    \centering
    \begin{minipage}[t]{0.49\textwidth}
        \vspace{0pt}
        \centering
        \includegraphics[width=\linewidth,height=0.67\linewidth,keepaspectratio]{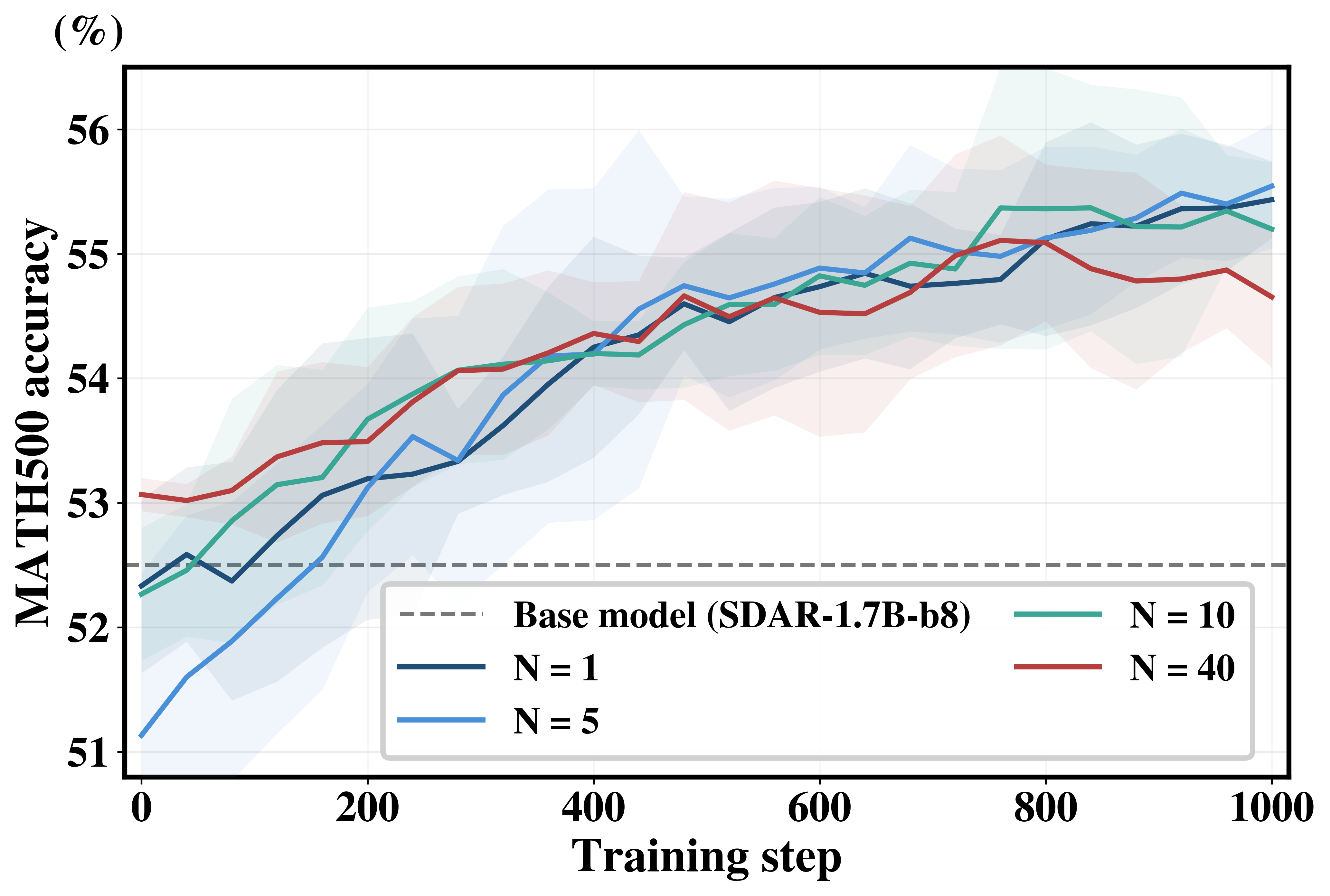}
        \caption{Training dynamics on MATH500 with different rollout reuse epochs $N$ on SDAR-1.7B-b8. Solid lines show accuracy trajectories; the dashed line denotes the base model.}
        \label{fig:rollout_replay_ablation}
    \end{minipage}
    \hfill
    \begin{minipage}[t]{0.49\textwidth}
        \vspace{0pt}
        \centering
        \includegraphics[width=\linewidth,height=0.67\linewidth,keepaspectratio]{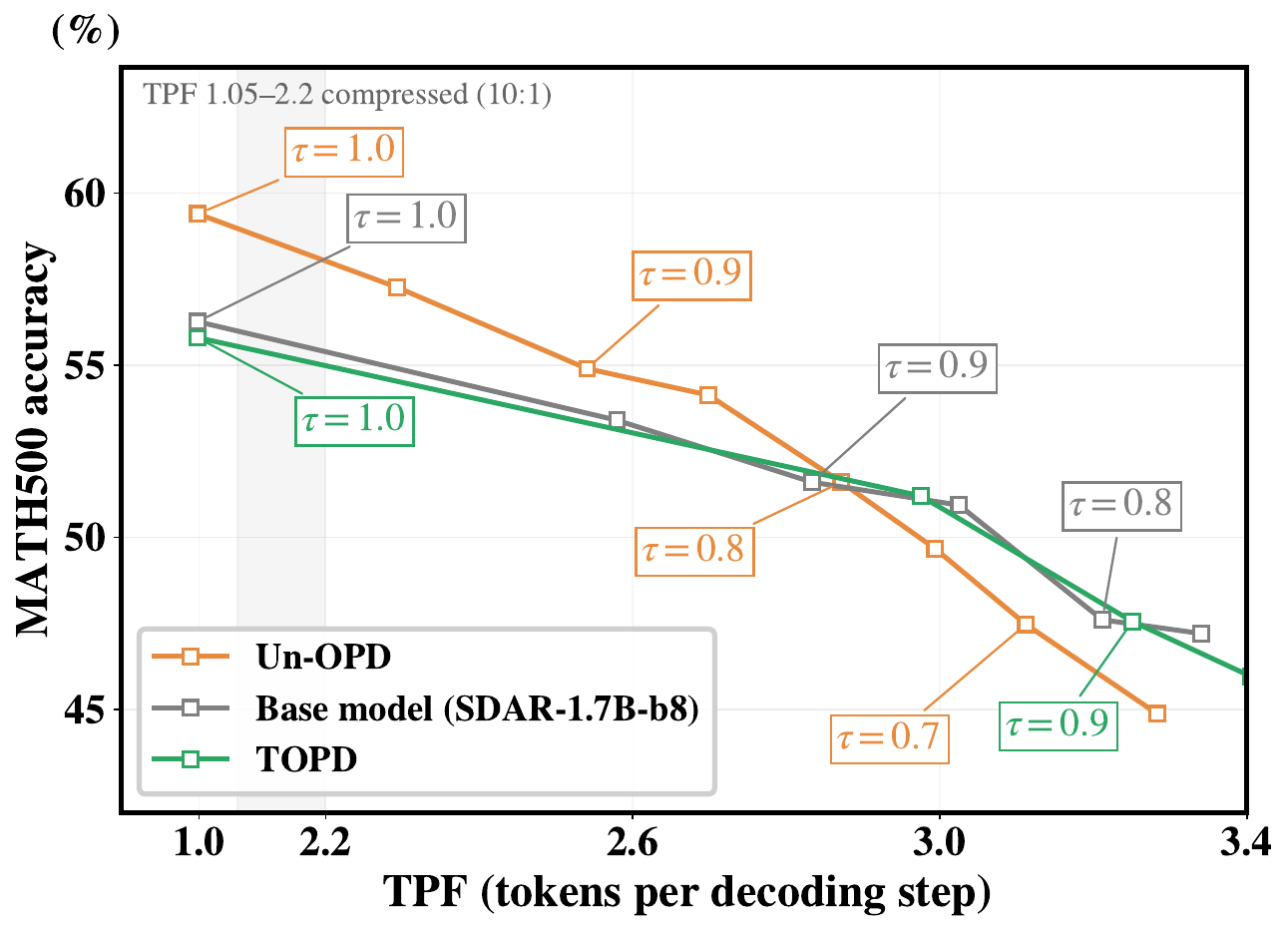}
        \caption{Accuracy versus tokens per forward pass (TPF) on MATH500 for Un-OPD and the SDAR-1.7B-b8 base model across decoding thresholds $\tau$.}
        \label{fig:tpf_accuracy}
    \end{minipage}
    \vspace{-2mm}
\end{figure}
For rollout reuse, Fig.~\ref{fig:rollout_replay_ablation} compares the training dynamics of \texttt{SDAR-1.7B-b8} under $N\in\{1,5,10,40\}$.
Moderate reuse largely preserves performance: by the end of training, $N=5$ and $N=10$ achieve similar performance to $N=1$ (the non-reused baseline).
In contrast, with $N=40$, performance declines during later training, likely due to off-policy shift from overly stale rollout trajectories.
Taken together, we set $\lambda=0.4$ and $N=10$ to balance accuracy and rollout generation cost.
%

\textbf{Analysis of tokens per forward pass (TPF).}
To evaluate the trade-off between reasoning accuracy and decoding parallelism, we sweep the dynamic decoding threshold $\tau$ on MATH500 (Fig.~\ref{fig:tpf_accuracy}). 
Across high-to-moderate confidence thresholds, \mbox{Un-OPD} consistently establishes a superior Pareto frontier over the base model, achieving up to a $+3.3\%$ accuracy gain under conservative decoding ($\tau=1.0$).
However, as decoding becomes overly aggressive (higher TPF), this performance advantage narrows and eventually diminishes, leaving room for further improvement.

\begin{wraptable}{r}{0.49\textwidth}
    \centering
    \vspace{-1mm}
    \caption{Cross-domain accuracy of SDAR-4B-b8. Arrows denote training $\rightarrow$ evaluation domains.}
    \vspace{-4mm}
    \label{tab:ood_capabilities}
    \setlength{\tabcolsep}{3pt}
    \renewcommand{\arraystretch}{1.1}
    \resizebox{\linewidth}{!}{%
    \begin{tabular}{@{}l|cc|cc@{}}
        \toprule
        \multirow{2}{*}{\textbf{Method}}
        & \multicolumn{2}{c|}{\textbf{Code $\rightarrow$ Math}}
        & \multicolumn{2}{c}{\textbf{Math $\rightarrow$ Code}} \\
        \cmidrule(lr){2-3}\cmidrule(l){4-5}
        & MATH500 & AMC2023
        & MBPP & HumanEval \\
        \midrule
        Base & 66.3 & 42.5 & 53.5 & 67.9 \\
        Un-OPD & 67.2 & 43.6 & 53.3 & 68.0 \\
        \bottomrule
    \end{tabular}%
    }
    \vspace{2mm}
\end{wraptable}
\textbf{Out-of-distribution (OOD) transferability.}
To test cross-domain robustness, we evaluate \texttt{SDAR-4B-b8} across math and code domains (Tab.~\ref{tab:ood_capabilities}).
After post-training on math, \mbox{Un-OPD} largely preserves code-generation performance, maintaining virtually identical accuracy on MBPP ($53.3\%$) and HumanEval ($68.0\%$).
Furthermore, code-trained \mbox{Un-OPD} exhibits modest positive transfer to math reasoning, outperforming Base by $+0.9\%$ on MATH500 and $+1.1\%$ on AMC2023.
These results confirm that \mbox{Un-OPD} avoids out-of-domain degradation while facilitating cross-task generalization.

\section{Conclusion}
\label{sec:conclusion}
In this work, we pioneer the study of scaling on-policy distillation (OPD) to block diffusion language models (BDLMs) with varying block sizes. 
We uncover two fundamental optimization bottlenecks: \textit{context misalignment} induced by mismatched block boundaries, and \textit{overconfidence collapse} triggered by biased learning dynamics under reverse-KL supervision. 
To overcome these challenges, we introduce \mbox{\textbf{Un-OPD}}, which eliminates supervisory distortions via boundary-aware step filtering and recalibrates optimization intensity across token supports. In addition, we integrate a rollout reuse mechanism that roughly halves training overhead with negligible accuracy loss. Comprehensive experiments on mathematical reasoning and code generation demonstrate that Un-OPD achieves remarkable stability and superior accuracy as well as substantial efficiency.
Despite these gains, the effectiveness of Un-OPD may depend on the quality and suitability of the teacher's supervision. Furthermore, pushing this Pareto frontier toward ultra-high parallelism regimes presents a compelling avenue for future research.

\subsection*{AI use statement}

We used OpenAI's ChatGPT/Codex tools to support the preparation and revision of this manuscript. This assistance included drafting and revising passages, improving grammar and wording, refining paragraph structure, and maintaining consistent terminology across sections. We also used these tools to discuss the presentation of methodological and theoretical explanations, including clarifying notation, making intermediate reasoning steps explicit, and improving the connection between the proposed method and its supporting analysis.
In the experimental discussion, the tools provided suggestions for experimental design and analysis and assisted with organizing comparisons and describing observed training dynamics. They also supported consistency checks across the main text, tables, figure captions, and appendices, as well as arithmetic calculations and numerical summaries based on experimental results supplied by the authors. Additional assistance involved LaTeX editing and formatting, including the presentation of equations, figures, tables, cross-references, and supplementary material.
The authors checked and reviewed all AI-generated content and results included in this work, including proposed wording, methodological and theoretical explanations, interpretations of experimental observations, and numerical calculations. The authors made the final decisions about which suggestions to incorporate and take responsibility for the complete final content of this work, including all text, scientific claims, analyses, and artifacts produced with AI assistance.

\subsection*{Ethics statement}
This work improves the training stability and efficiency of block diffusion language models through on-policy distillation. Students trained with Un-OPD may inherit biases, factual errors, or unsafe code-generation behaviors from the teacher and training data. The reported gains on mathematical reasoning and code-generation benchmarks do not establish safety or reliability beyond the evaluated settings. More efficient generation may also lower the cost of producing misleading content or insecure code at scale. Applications of the method should therefore include task-specific safety evaluation and appropriate review of generated outputs.

\subsection*{Reproducibility statement}
To support reproducibility, Sec.~\ref{sec:method} specifies the Un-OPD objective, context-aligned step selection, support-rebalanced calibration, and rollout reuse mechanism, while Algorithm~\ref{alg:un_opd} in Appendix~\ref{app:algorithm_details} presents the training procedure. Experimental settings, model configurations, and data sources are described in Sec.~\ref{exp_setting} and Appendix~\ref{app:experimental_details}, which also documents hardware, training hyperparameters, prompt templates, baseline configurations, decoding settings, evaluation metrics, and answer-verification procedures. For the theoretical result, Appendix~\ref{app:proof_top1_logit_asymmetry} provides the complete proof of Proposition~\ref{prop:agreement_head_gradient}, together with its assumptions and scope. Additional training-dynamics analyses and the procedures used to construct the corresponding smoothed curves are provided in Appendix~\ref{app:further_experimental_analysis}.

\bibliography{iclr2027_conference}
\bibliographystyle{iclr2027_conference}

\ifincludeappendix
    \clearpage
    \appendix
\section{Framework Details}
\label{app:framework_details}
\subsection{Algorithm Details}
\label{app:algorithm_details}

Algorithm~\ref{alg:un_opd} summarizes the Un-OPD training loop. Following TOPD~\cite{ren2026trace}, each rollout round comprises three stages: sampling prompts and responses from the current student, constructing training examples from the recorded denoising traces, and optimizing the student with frozen-teacher supervision. 
Each sample pairs a committed token position with its decoding state, attaching supervision to decisions made along the student's actual generation process. We reconstruct the visible context under each model's block schedule and retain only decisions satisfying $s_{\mathrm{T},k}=s_{\mathrm{S},k}$, so teacher and student predictions are compared under the same context. 
At retained positions with $\rho_{k,i}>1$, teacher logits are interpolated with detached student logits before computing the reverse KL. The resulting trace batch is reused for $N$ optimization epochs to reduce rollout sampling overhead.

\begin{algorithm}[H]
    \small
    \DontPrintSemicolon
    \LinesNumbered
    \SetKwInput{KwRequire}{Require}
    \SetKwFor{ForEach}{for each}{do}{end}
    \caption{Un-OPD: Context-Aligned Distillation with Support Rebalance}
    \label{alg:un_opd}
    \KwRequire{Student $\pi_\theta$; frozen teacher $\pi_{\mathrm{T}}$; prompt set $\mathcal{D}$; prompt batch size $M$; rollouts per prompt $G$; update minibatch size $m$ (traces); student/teacher block sizes $B_{\mathrm{S}},B_{\mathrm{T}}$; student decoding schedule; maximum response length $L_{\max}$; calibration coefficient $\lambda$; replay epochs $N$; learning rate $\eta$.}
    \For{each rollout round}{
        Sample a prompt batch $\mathcal{B}\sim\mathcal{D}$ with $|\mathcal{B}|=M$; initialize buffer $\mathcal{R}\leftarrow\varnothing$\;
        \tcp{Collect student traces and select context-aligned decisions}
        \ForEach{$q\in\mathcal{B}$ and $g=1,\ldots,G$}{
            Initialize masked response state $s_0$; set $k\leftarrow0$\;
            \While{the response is unfinished and below $L_{\max}$}{
                Sample committed decisions $a_k\sim\pi_\theta(\cdot\mid q,s_k)$ under the decoding schedule\;
                $s_{k+1}\leftarrow\operatorname{unmask}(s_k,a_k)$; $k\leftarrow k+1$\;
            }
            Store trace $\xi=(s_0,a_0,\ldots,a_{K-1},s_K)$ with $K=k$\;
            Reconstruct visible states $s_{\mathrm{S},k},s_{\mathrm{T},k}$ using $B_{\mathrm{S}},B_{\mathrm{T}}$\;
            $\mathcal{A}_{\mathrm{align}}(\xi)\leftarrow\{(k,i):(i,x_i)\in a_k,\ 0\le k<K,\ s_{\mathrm{T},k}=s_{\mathrm{S},k}\}$\;
            \If{$\mathcal{A}_{\mathrm{align}}(\xi)\neq\varnothing$}{
                Add $(q,\xi,\mathcal{A}_{\mathrm{align}}(\xi))$ to $\mathcal{R}$\;
            }
        }
        \If{$\mathcal{R}=\varnothing$}{
            Repeat rollout sampling and alignment filtering until $\mathcal{R}\neq\varnothing$\;
        }
        \tcp{Replay fixed traces with support-aware teacher supervision}
        \For{$n=1,\ldots,N$}{
            Shuffle $\mathcal{R}$ and split it into minibatches of at most $m$ traces\;
            \ForEach{minibatch $\mathcal{M}$}{
                \ForEach{$(q,\xi,\mathcal{A}_{\mathrm{align}}(\xi))\in\mathcal{M}$}{
                    \ForEach{$(k,i)\in\mathcal{A}_{\mathrm{align}}(\xi)$}{
                        Compute student logits $z_{\mathrm{S},k,i}$ at $(q,s_{\mathrm{S},k},i)$ and frozen teacher logits $z_{\mathrm{T},k,i}$ at $(q,s_{\mathrm{T},k},i)$\;
                        $p_{\mathrm{S},k,i}\leftarrow\operatorname{softmax}(z_{\mathrm{S},k,i})$; $p_{\mathrm{T},k,i}\leftarrow\operatorname{softmax}(z_{\mathrm{T},k,i})$\;
                        $h_{k,i}\leftarrow\arg\max_v p_{\mathrm{S},k,i}(v)$; $\rho_{k,i}\leftarrow p_{\mathrm{T},k,i}(h_{k,i})/p_{\mathrm{S},k,i}(h_{k,i})$\;
                        $\widetilde z_{\mathrm{T},k,i}\leftarrow z_{\mathrm{T},k,i}$\;
                        \If{$\rho_{k,i}>1$}{
                            $\widetilde z_{\mathrm{T},k,i}\leftarrow\lambda z_{\mathrm{T},k,i}+(1-\lambda)\operatorname{sg}[z_{\mathrm{S},k,i}]$\;
                        }
                        $\ell_{k,i}\leftarrow D_{\mathrm{KL}}\!\left(p_{\mathrm{S},k,i}\,\Vert\,\operatorname{sg}[\operatorname{softmax}(\widetilde z_{\mathrm{T},k,i})]\right)$\;
                    }
                    $\mathcal{L}_{\xi}\leftarrow |\mathcal{A}_{\mathrm{align}}(\xi)|^{-1}\sum_{(k,i)\in\mathcal{A}_{\mathrm{align}}(\xi)}\ell_{k,i}$\;
                }
                $\mathcal{L}_{\mathcal{M}}\leftarrow|\mathcal{M}|^{-1}\sum_{\xi\in\mathcal{M}}\mathcal{L}_{\xi}$; $\theta\leftarrow\theta-\eta\nabla_\theta\mathcal{L}_{\mathcal{M}}$\;
            }
        }
    }
\end{algorithm}

Rollout states and committed decisions are held fixed during replay, while student predictions, top-1 tokens, and calibration targets are recomputed at each update. No gradient passes through the teacher, the support-based selection, or the calibrated targets. Traces without aligned decisions are filtered. If all traces in a batch are filtered out, we repeat rollout sampling and alignment filtering until the buffer is nonempty, then proceed with student updates. Averaging within each trace before averaging across traces follows Eq.~\ref{eq:un_opd_objective}; the parameter update is written as gradient descent for clarity.

\section{Experimental Details}
\label{app:experimental_details}

This section documents the training configuration, prompt templates, baseline settings, and evaluation benchmarks for our experiments on \texttt{SDAR} series models.

\subsection{Training Configuration}
\label{app:training_configuration}

Tab.~\ref{tab:training_configuration} summarizes the detailed experimental setup for \mbox{Un-OPD}, covering model architectures, rollout sampling, optimization, and datasets. Overall configurations follow TOPD~\cite{ren2026trace}, with key hyperparameters for context-aligned step selection, support rebalance ($\lambda$), and rollout replay ($N$) explicitly detailed. The paired model scales and block sizes are consistent with those in Tab.~\ref{tab:math_and_code}, while the specific effects of $\lambda$ and $N$ are thoroughly investigated in Sec.~\ref{ablation}.

\begin{table}[H]
    \centering
    \small
    \setlength{\tabcolsep}{5pt}
    \renewcommand{\arraystretch}{1.04}
    \caption{Training configuration for Un-OPD on SDAR students.}
    \label{tab:training_configuration}
    \begin{tabular}{@{}p{0.43\linewidth}p{0.53\linewidth}@{}}
        \toprule
        \textbf{Category} & \textbf{Value} \\
        \midrule
        \multicolumn{2}{@{}l}{\textit{Model and Initialization}} \\
        Student models & \texttt{SDAR-1.7B} and 
        \texttt{SDAR-4B} \\
        Teacher model & \texttt{TraDo-8B} \\
        Hardware configuration & Single node with 8 NVIDIA H20 GPUs \\
        Precision & bfloat16; TF32 enabled \\
        Gradient checkpointing & Enabled (whole layer) \\
        \midrule
        \multicolumn{2}{@{}l}{\textit{Rollout (On-Policy Sampling)}} \\
        Tasks (prompts) per rollout round & 64 \\
        Responses per prompt & 1 \\
        Sampling temperature & 1.0 \\
        Maximum generation length & 2,000 tokens \\
        Student / teacher block sizes & Per configuration in Tab.~\ref{tab:math_and_code} \\
        Denoising steps per block & student block size \\
        Remasking strategy & Low-confidence dynamic ($\tau=0.9$) \\
        Top-$p$ & 1.0 \\
        \midrule
        \multicolumn{2}{@{}l}{\textit{Training and Optimization}} \\
        Optimizer & AdamW \\
        Learning rate & $2\times10^{-7}$ (constant) \\
        Weight decay & 0.0 \\
        Effective optimization batch size & $32$ (4/GPU $\times$ 8 $\times$ 1 accum.) \\
        Gradient clipping (max norm) & $1.0$ \\
        \midrule
        \multicolumn{2}{@{}l}{\textit{Un-OPD-Specific Settings}} \\
        Loss type & Full-vocabulary reverse KL \\
        Teacher status & Frozen \\
        Interpolation coefficient $\lambda$ & $0.4$ \\
        Rollout replay epochs $N$ & $10$ \\
        \midrule
        \multicolumn{2}{@{}l}{\textit{Data}} \\
        Training dataset & 8K math problems following TraceRL~\cite{wang2026revolutionizing} \\
        Maximum prompt length & 784 tokens \\
        \bottomrule
    \end{tabular}
\end{table}

\textbf{Training infrastructure and execution.}
We train all \mbox{Un-OPD} models on a single node equipped with eight NVIDIA H20 GPUs. Each rollout cycle samples a batch of 64 prompts, followed by $N$ optimization epochs over the retained, context-aligned traces. During rollout replay, the teacher parameters and rollout trajectories remain frozen, whereas student predictions and the resulting calibrated targets are dynamically recomputed at each gradient update, as detailed in Algorithm~\ref{alg:un_opd}.

\subsection{Prompt Template}
\label{app:prompt_template}
Fig.~\ref{fig:sdar_math_prompt} presents the math prompt template for \texttt{SDAR} models. The placeholder \texttt{\{\{problem\}\}} is replaced with the problem statement. The instruction requests step-by-step reasoning and a final answer enclosed in \texttt{\textbackslash boxed\{\}}, followed by the assistant prefix from which generation begins.

\begin{figure}[H]
    \centering
    \begin{tcolorbox}[
        width=0.95\linewidth,
        colback=black!2,
        colframe=black!30,
        colbacktitle=black!30,
        coltitle=white,
        title={Math prompt template},
        fonttitle=\large\bfseries,
        boxrule=0.8pt,
        arc=3mm,
        left=12pt, right=12pt, top=9pt, bottom=9pt,
        before skip=0pt, after skip=0pt
    ]
        \textbf{\texttt{SDAR}}\par
        {\small\ttfamily\raggedright\setlength{\parindent}{0pt}
        \verb!<|im_start|>user!\par
        \verb!{{problem}}!\par
        Please reason step by step, and put your final answer within \verb!\boxed{}.<|im_end|>!\par
        \verb!<|im_start|>assistant!\par
        }
    \end{tcolorbox}
    \vspace{4mm}
    \caption{Math prompt template for SDAR, with \texttt{\{\{problem\}\}} denoting the input problem.}
    \label{fig:sdar_math_prompt}
\end{figure}

For code generation, Fig.~\ref{fig:sdar_code_prompt} shows the template with an in-context prefix, \texttt{\{\{icl\_prefix\}\}}, prepended to the problem statement. The instruction requires the response code to appear in a single Python code block.

\begin{figure}[H]
    \centering
    \begin{tcolorbox}[
        width=0.95\linewidth,
        colback=black!2,
        colframe=black!30,
        colbacktitle=black!30,
        coltitle=white,
        title={Code prompt template},
        fonttitle=\large\bfseries,
        boxrule=0.8pt,
        arc=3mm,
        left=12pt, right=12pt, top=9pt, bottom=9pt,
        before skip=0pt, after skip=0pt
    ]
        \textbf{\texttt{SDAR}}\par
        {\small\ttfamily\raggedright\setlength{\parindent}{0pt}
        \verb!<|im_start|>user!\par
        \verb!{{icl_prefix}}{{problem}}!\par
        Place your code within a single Python code block \verb!`python `!. Do not include more than one code block. \verb!<|im_end|>!\par
        \verb!<|im_start|>assistant!\par
        }
    \end{tcolorbox}
    \vspace{4mm}
    \caption{Code prompt template for SDAR. The placeholders denote the in-context prefix and the input problem.}
    \label{fig:sdar_code_prompt}
\end{figure}

\subsection{Baseline Configurations}
\label{app:baseline_configurations}
We compare Un-OPD with trace-based reinforcement learning (\textbf{TraceRL}~\cite{wang2026revolutionizing}), supervised fine-tuning, and trace-aligned on-policy distillation (\textbf{TOPD}). 
Student scales and block sizes follow the corresponding groups in Tab.~\ref{tab:math_and_code}.

\textbf{TraceRL.}
We train each student model using the TraceRL codebase~\cite{wang2026revolutionizing}.
TraceRL combines the denoising-trace formulation with group-relative policy optimization~\cite{shao2024deepseekmath}. For each prompt, it samples a group of $G=16$ responses and computes relative advantages from their binary correctness rewards. Each complete response receives a reward of $1$ if it solves the problem correctly and $0$ otherwise.
The update uses token log-probabilities evaluated at the recorded decoding states, with old-policy log-probabilities held fixed throughout optimization of the rollout batch. A clipped policy objective limits policy changes, while a KL penalty regularizes the student toward a frozen reference policy. We use a clipping threshold of $\epsilon=0.2$, a KL coefficient of $\beta=0.01$, a learning rate of $1\times10^{-6}$ with cosine decay, and an effective optimization batch size of $16$. Advantages are computed within each response group, without a learned value model for simplicity.

\textbf{Semi-autoregressive SFT.}
The SFT baseline fine-tunes the student on offline teacher trajectories using the block-diffusion masking objective established by \texttt{SDAR}~\cite{cheng2026sdar} and TraceRL~\cite{wang2026revolutionizing}.
Specifically, fully generated preceding blocks serve as causal context, while training optimizes the cross-entropy loss over masked tokens within the supervised block.
Following the SFT recipe in TraceRL, the masking ratio is drawn uniformly from $[0.1, 0.9]$, and the block size is configured to match the student under evaluation.
Models are trained for one epoch with a learning rate of $1\times10^{-5}$. 
For mathematical reasoning, the training corpus consists of dynamic decoding responses sampled by the \texttt{TraDo-8B} teacher on the same 8K math problems, following the training setup of TraceRL~\cite{wang2026revolutionizing}.
Likewise, code training trajectories are sampled from the AceCoder subset using the same teacher model.

\textbf{TOPD.}
TOPD~\cite{ren2026trace} serves as the standard trace-aligned on-policy distillation baseline. It samples rollouts from the student policy and supervises committed token decisions along intermediate denoising traces. At each supervised position, it minimizes the full-vocabulary reverse-KL divergence against a frozen teacher evaluated directly on the student's decoding states. 
TOPD directly applies standard OPD to BDLMs without accounting for how supervision at different denoising steps affects student optimization.
For a fair comparison, we reproduce TOPD across different student models using the configurations reported in the original paper.
%
%
\subsection{Evaluation Benchmarks}
\label{app:evaluation_benchmarks}
We evaluate mathematical reasoning and code generation across six widely adopted benchmarks, as summarized in Tab.~\ref{tab:math_and_code}.

\noindent\textbf{Mathematical reasoning.}
We assess problem-solving capabilities across a diverse spectrum of difficulty using four benchmarks: 
(1) \textbf{GSM8K} evaluates multi-step grade-school arithmetic reasoning; 
(2) \textbf{MATH-500} comprises a 500-problem subset of the challenging MATH benchmark across algebra, geometry, and number theory; 
(3) \textbf{AMC 2023} comprises 40 competition-level problems drawn from the 2023 American Mathematics Competitions; and 
(4) \textbf{OlympiadBench} tests advanced, Olympiad-level mathematical reasoning. Together, these benchmarks span elementary arithmetic to elite competition mathematics.

\noindent\textbf{Code generation.}
We employ \textbf{HumanEval} and \textbf{MBPP} to evaluate standalone Python program synthesis from docstrings and natural-language problem descriptions. For both benchmarks, correctness is strictly assessed through the functional execution of official unit tests.

\noindent\textbf{Sampling and evaluation metrics.}
We evaluate BDLMs using a dynamic decoding schedule (confidence threshold $\tau=0.9$), with a temperature of $1.0$ and top-$p=1.0$, under their respective block sizes unless otherwise specified.
We report the average accuracy metric, $\operatorname{avg}@k$, computed as the mean instance-level pass rate over $k$ sampled responses per problem. We sample four responses per problem and report $\operatorname{avg}@4$ for all benchmarks except AMC 2023, for which we sample 16 responses per problem and report $\operatorname{avg}@16$ to mitigate sampling variance arising from its compact 40-problem test set.

\noindent\textbf{Answer verification.}
For all mathematical reasoning benchmarks (GSM8K, MATH-500, AMC 2023, and OlympiadBench), we extract the final answer from \texttt{\textbackslash boxed\{\}} and verify its mathematical equivalence to the reference answer via symbolic and numerical normalization. For MBPP and HumanEval, we execute the extracted code directly in a sandboxed Python environment, counting a response as correct only if it passes all assertion test cases.


\section{Further Experimental Analysis}
\label{app:further_experimental_analysis}


\subsection{Comparison under Different Context Alignment Regimes}
\label{app:context_alignment_regimes}
Here we investigate how the different context regimes defined in Sec.~\ref{sec:context_aligned_distillation} affect student model performance. 
Complementing the teacher-support and confidence analysis for \texttt{SDAR-4B-b8} in Fig.~\ref{fig:context_misalignment_analysis}, we train separate OPD runs using only the decoding steps from each regime: $G_1$, where the student sees more context than the teacher; $G_2$, where the teacher sees more context than the student; and $G_3$, where their visible contexts are aligned. Fig.~\ref{fig:context_regime_training_dynamics} compares their MATH500 accuracy under static and dynamic decoding.
In Fig.~\ref{fig:context_regime_training_dynamics}, scatter points denote raw evaluation accuracies at individual checkpoints, while solid curves trace exponential moving averages (EMA, $\alpha=0.35$). The surrounding shaded bands represent $\pm1$ local standard deviation computed over a backward sliding window of size five (including the current and up to four prior checkpoints). 
This identical visualization scheme is consistently adopted in Figs.~\ref{fig:static_eval_dynamics}, \ref{fig:matched_block_training_dynamics}, and~\ref{fig:larger_block_training_dynamics}.
\begin{figure}[!t]
    \centering
    \begin{subfigure}[b]{0.49\textwidth}
        \centering
        \includegraphics[width=\linewidth]{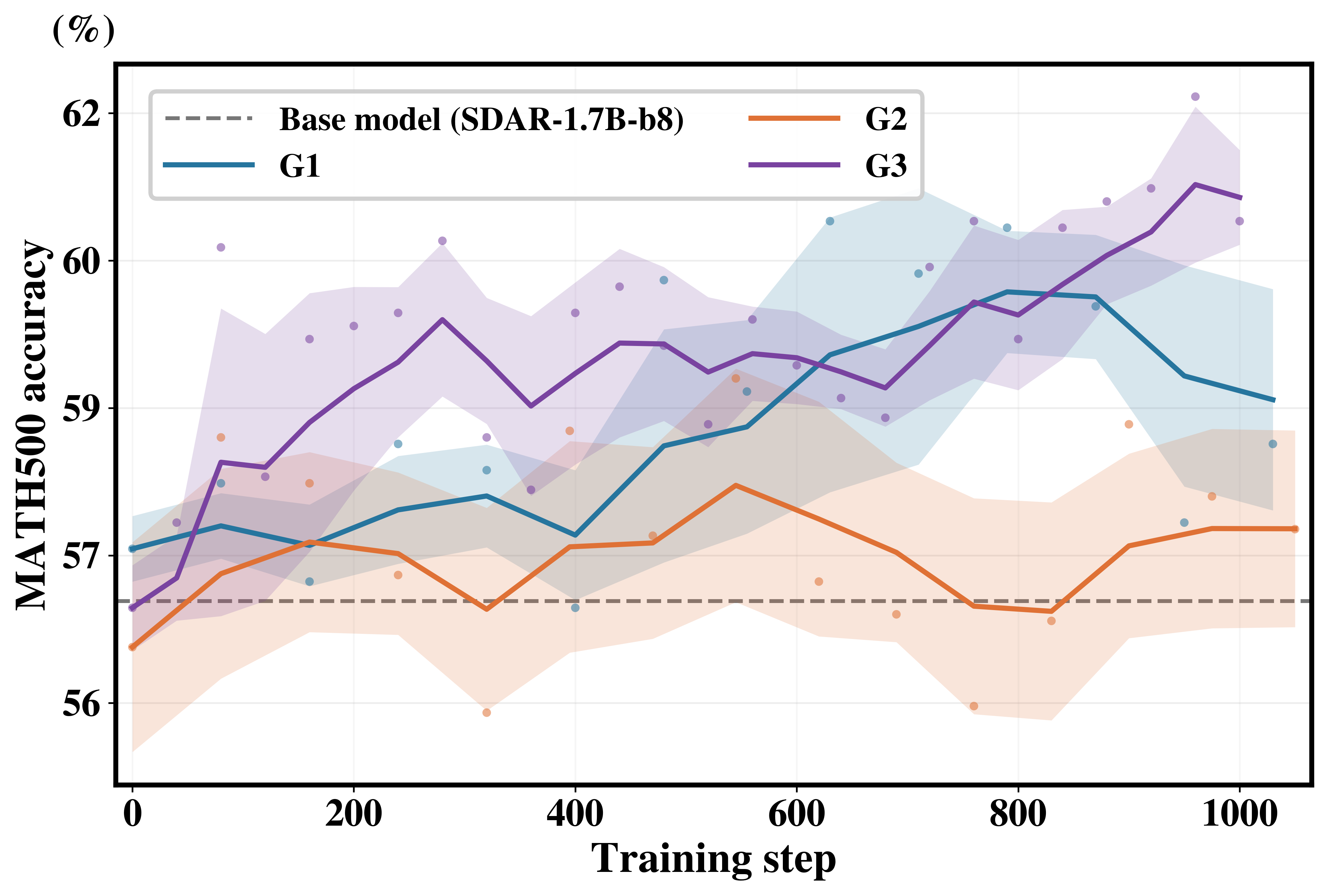}
        \caption{Static decoding.}
        \label{fig:context_regime_static}
    \end{subfigure}
    \hfill
    \begin{subfigure}[b]{0.49\textwidth}
        \centering
        \includegraphics[width=\linewidth]{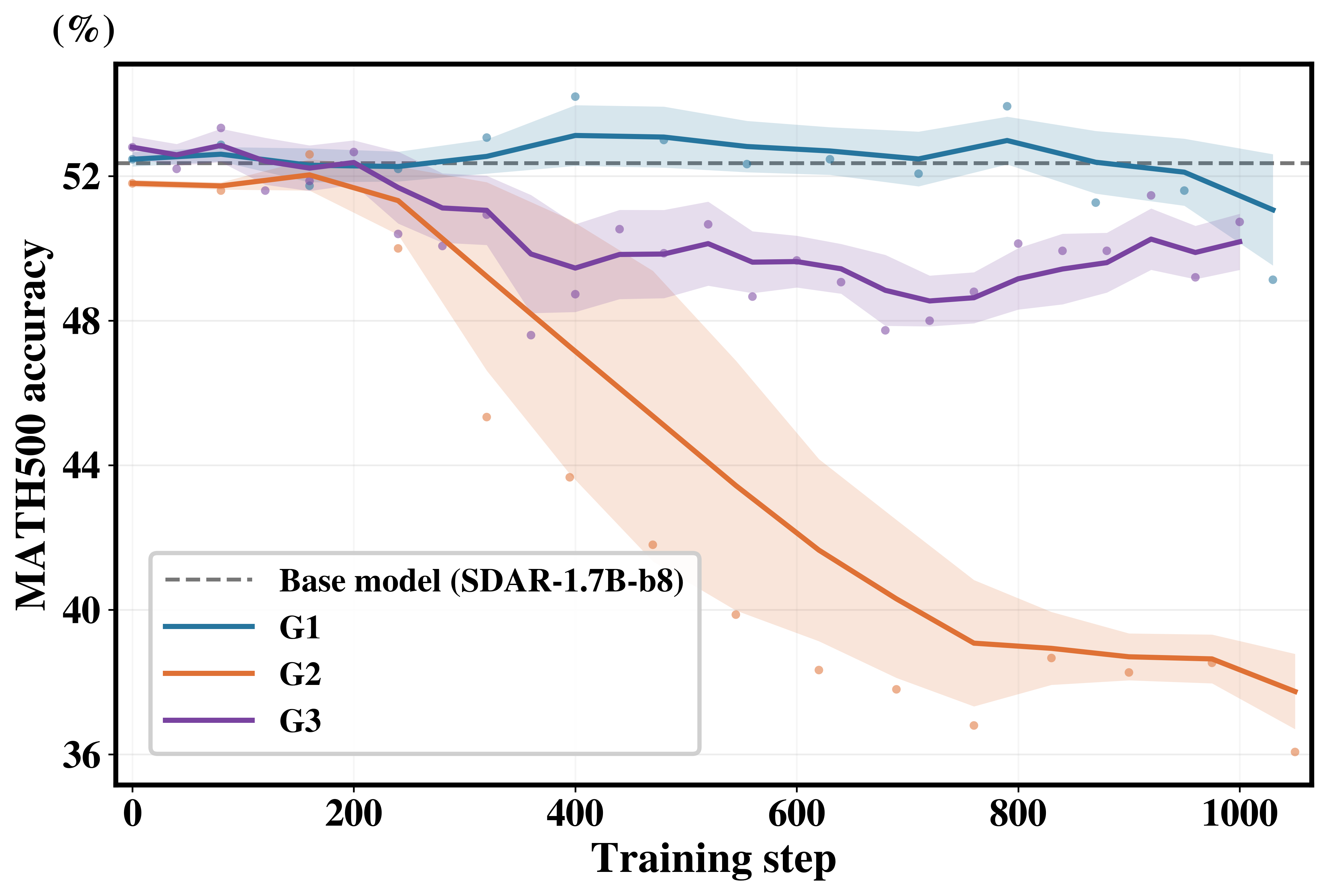}
        \caption{Dynamic decoding.}
        \label{fig:context_regime_dynamic}
    \end{subfigure}
    \caption{
        \textbf{Training dynamics under different context alignment regimes.}
        MATH500 accuracy for \texttt{SDAR-1.7B-b8} trained with OPD restricted to $G_1$ (blue), $G_2$ (orange), or $G_3$ (purple), as defined in Sec.~\ref{sec:context_aligned_distillation}. Points show checkpoint accuracies and solid curves show smoothed trends. Gray dashed lines indicate the base-model accuracy under each decoding schedule.
    }
    \label{fig:context_regime_training_dynamics}
\end{figure}

In Fig.~\ref{fig:context_regime_training_dynamics}(\subref{fig:context_regime_static}), training on $G_3$ obtains the strongest distillation performance, with the smoothed accuracy reaching approximately $61\%$. The $G_1$ run also improves over the base model, but loses part of its gains near the end of training. 
We attribute this to the lower teacher support inherent to $G_1$ steps, which penalizes confident student predictions even when correct, thereby inducing underconfidence and curtailing the model's ultimate decoding potential.
Meanwhile, training solely on $G_2$ achieves little gain over the baseline throughout training. We argue that $G_2$ steps inject miscalibrated, overly aggressive supervision, rewarding high confidence on inaccurate predictions and consequently degrading the model's true decoding capacity.
These results is consistent with the teacher signals biases identified in Sec.~\ref{sec:context_aligned_distillation}: aligned contexts provide more balanced teacher support, whereas either direction of context mismatch can reduce the effectiveness of distillation.
In Fig.~\ref{fig:context_regime_training_dynamics}(\subref{fig:context_regime_dynamic}), we find that $G_1$ marginally outperforms the other two regimes. We attribute this to the student's lower confidence under $G_1$, which favors dynamic decoding by encouraging overly cautious unmasking schedules. However, this superficial boost comes at the cost of the model's fundamental decoding capacity, as demonstrated by its degraded static performance. Consequently, in our context-aligned distillation, we intentionally restrict OPD to aligned steps ($G_3$) to preserve the model's performance.

\begin{figure}[H]
    \centering
    \begin{subfigure}[t]{0.49\textwidth}
        \centering
        \includegraphics[width=\linewidth]{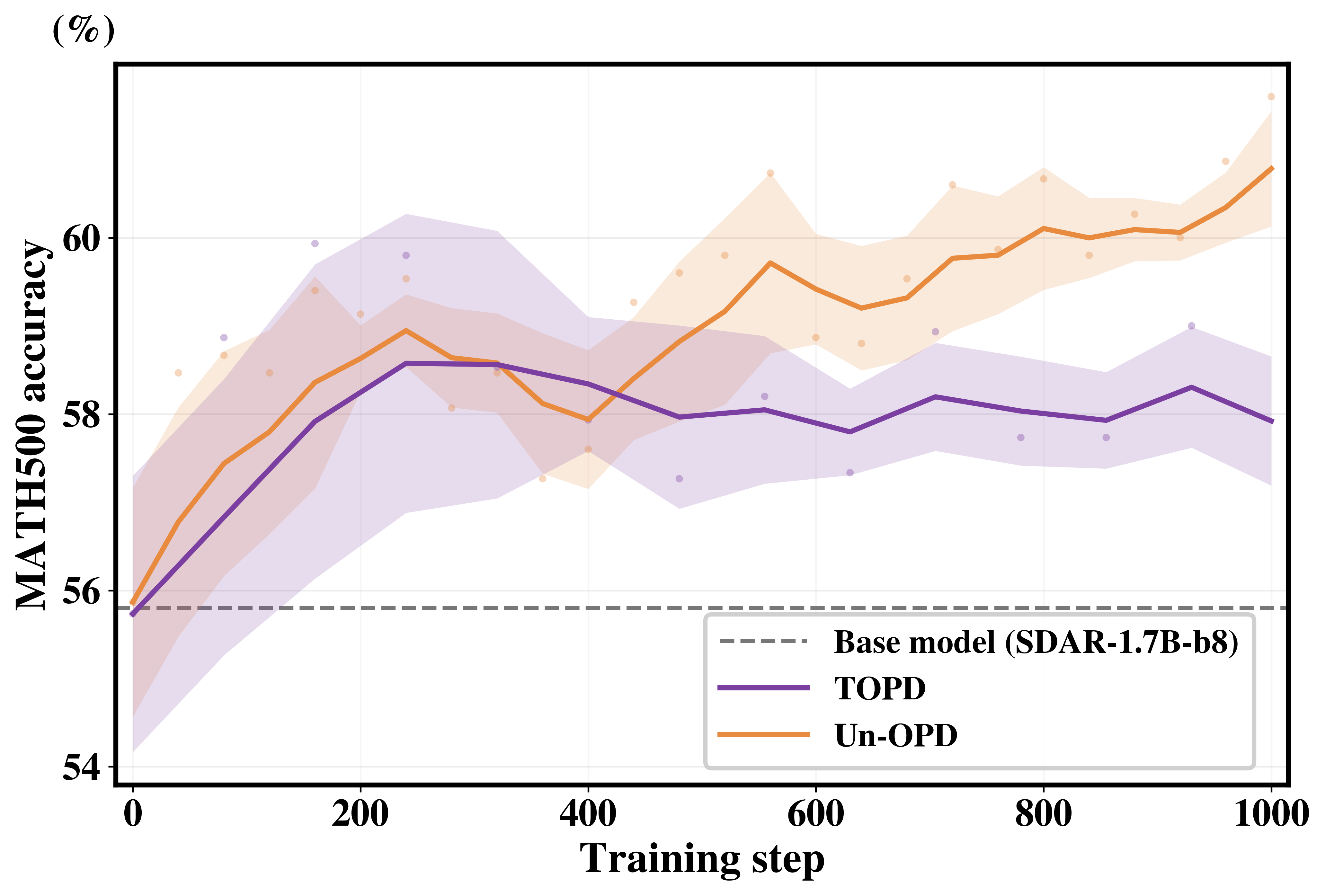}
        \caption{SDAR-1.7B-b8.}
        \label{fig:static_eval_dynamics_1p7b}
    \end{subfigure}
    \hfill
    \begin{subfigure}[t]{0.49\textwidth}
        \centering
        \includegraphics[width=\linewidth]{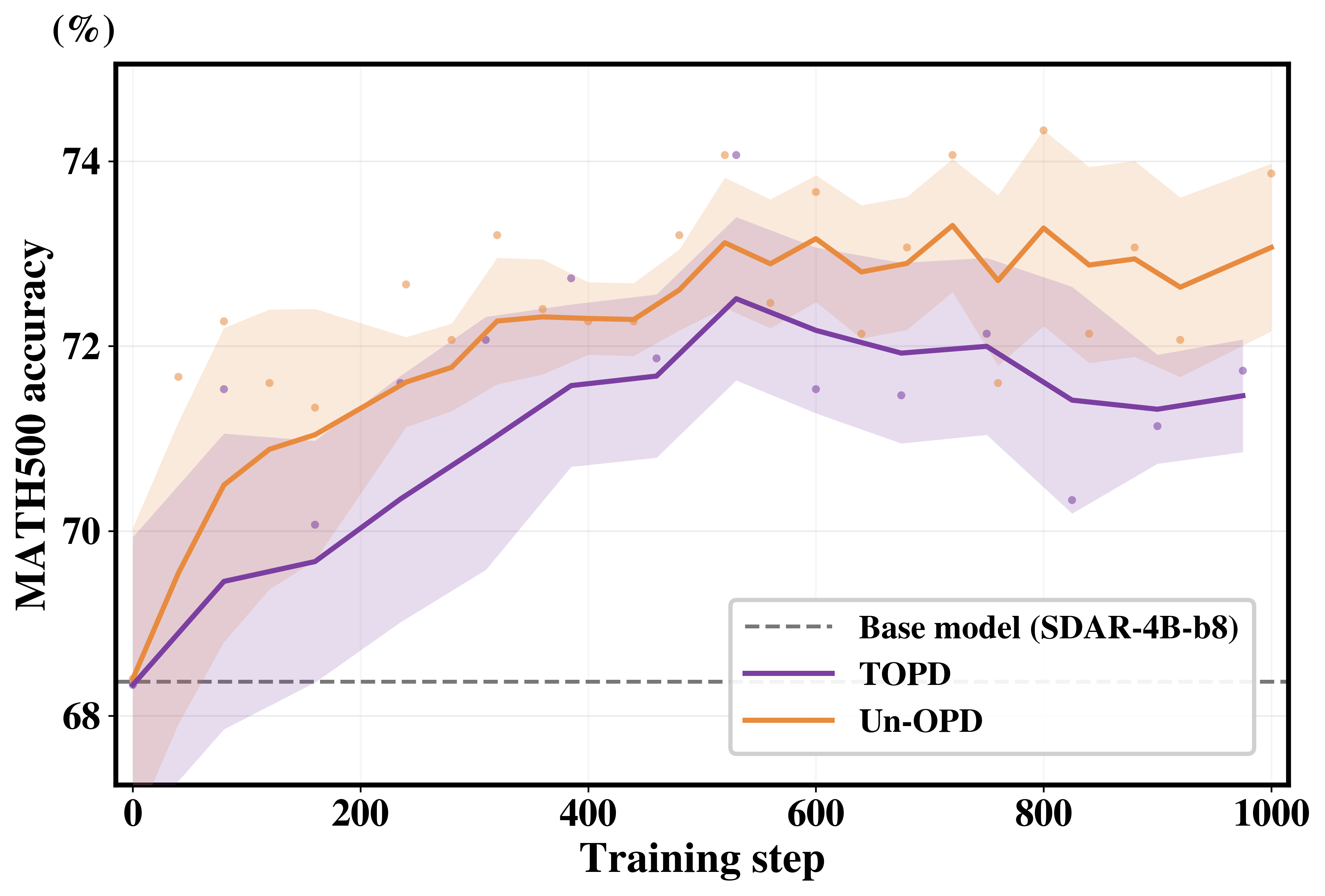}
        \caption{SDAR-4B-b8.}
        \label{fig:static_eval_dynamics_4b}
    \end{subfigure}
    \caption{
        \textbf{Training dynamics on MATH500 under static decoding.}
        Purple and orange curves correspond to TOPD and Un-OPD, respectively. 
    }
    \label{fig:static_eval_dynamics}
\end{figure}

\subsection{Comparison under a Static Decoding Schedule}
\label{app:static_decoding_schedule}

Different from the dynamic decoding schedule, the static schedule unmasks exactly one token per step, serving as an indicator of intrinsic model capability.
To complement the dynamic-decoding results in Fig.~\ref{fig:dynamic_eval_ablation}, we evaluate TOPD and Un-OPD under static decoding throughout training (Fig.~\ref{fig:static_eval_dynamics}).
For \texttt{SDAR-1.7B-b8}, both methods improve early on, but TOPD quickly plateaus around $58\%$, whereas Un-OPD continues climbing to $60.8\%$ at step 1,000 ($+2.9\%$ over TOPD). 
For \texttt{SDAR-4B-b8}, Un-OPD stabilizes at approximately $73.0\%$ around step 500, while TOPD degrades after peaking.
Collectively, these results confirm that Un-OPD not only excels under dynamic decoding schedule, but also substantially elevates the model's intrinsic performance upper bound.

\begin{figure}[t]
    \centering
    \begin{subfigure}[t]{0.49\textwidth}
        \centering
        \includegraphics[width=\linewidth]{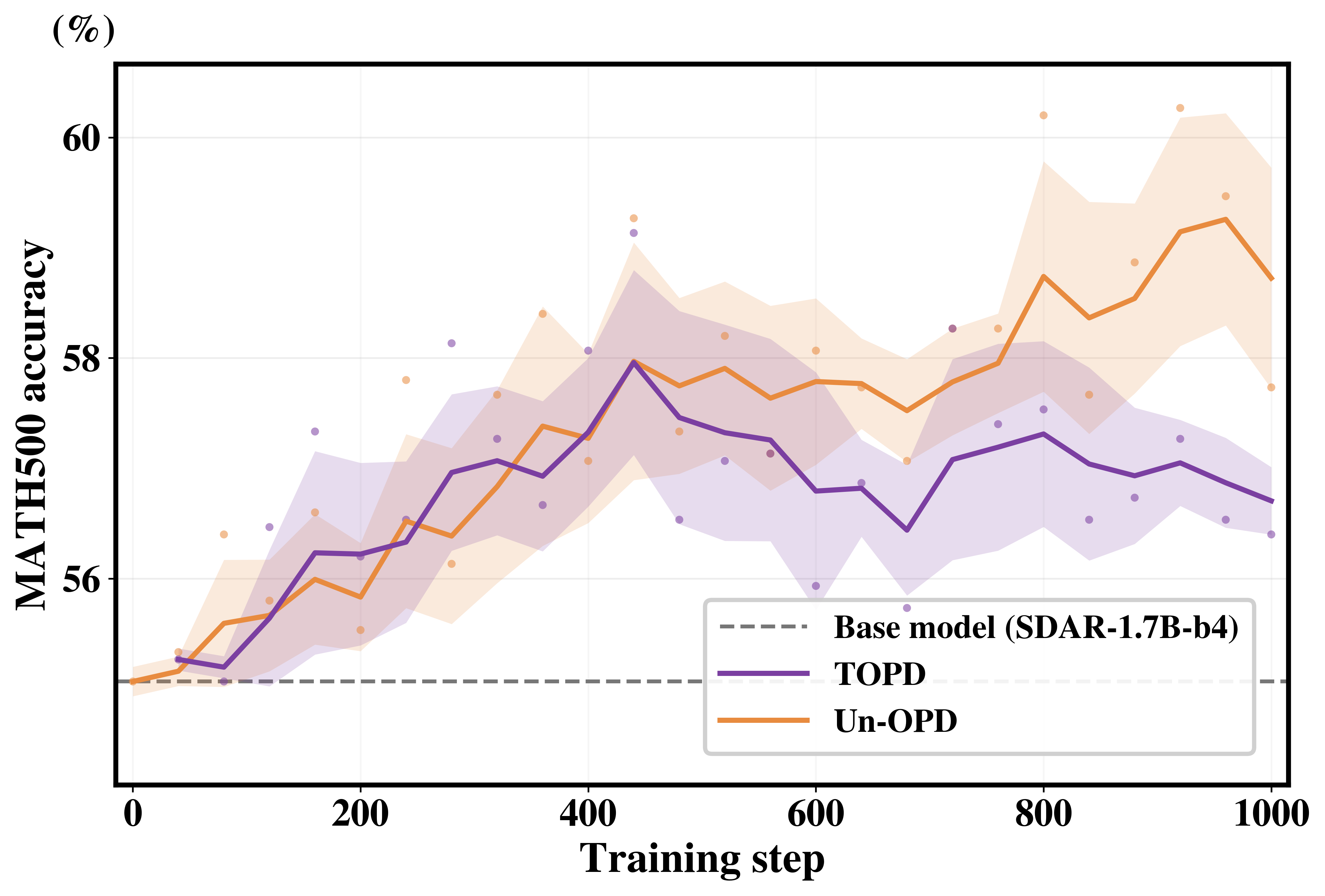}
        \caption{SDAR-1.7B-b4.}
        \label{fig:matched_block_training_dynamics_1p7b}
    \end{subfigure}
    \hfill
    \begin{subfigure}[t]{0.49\textwidth}
        \centering
        \includegraphics[width=\linewidth]{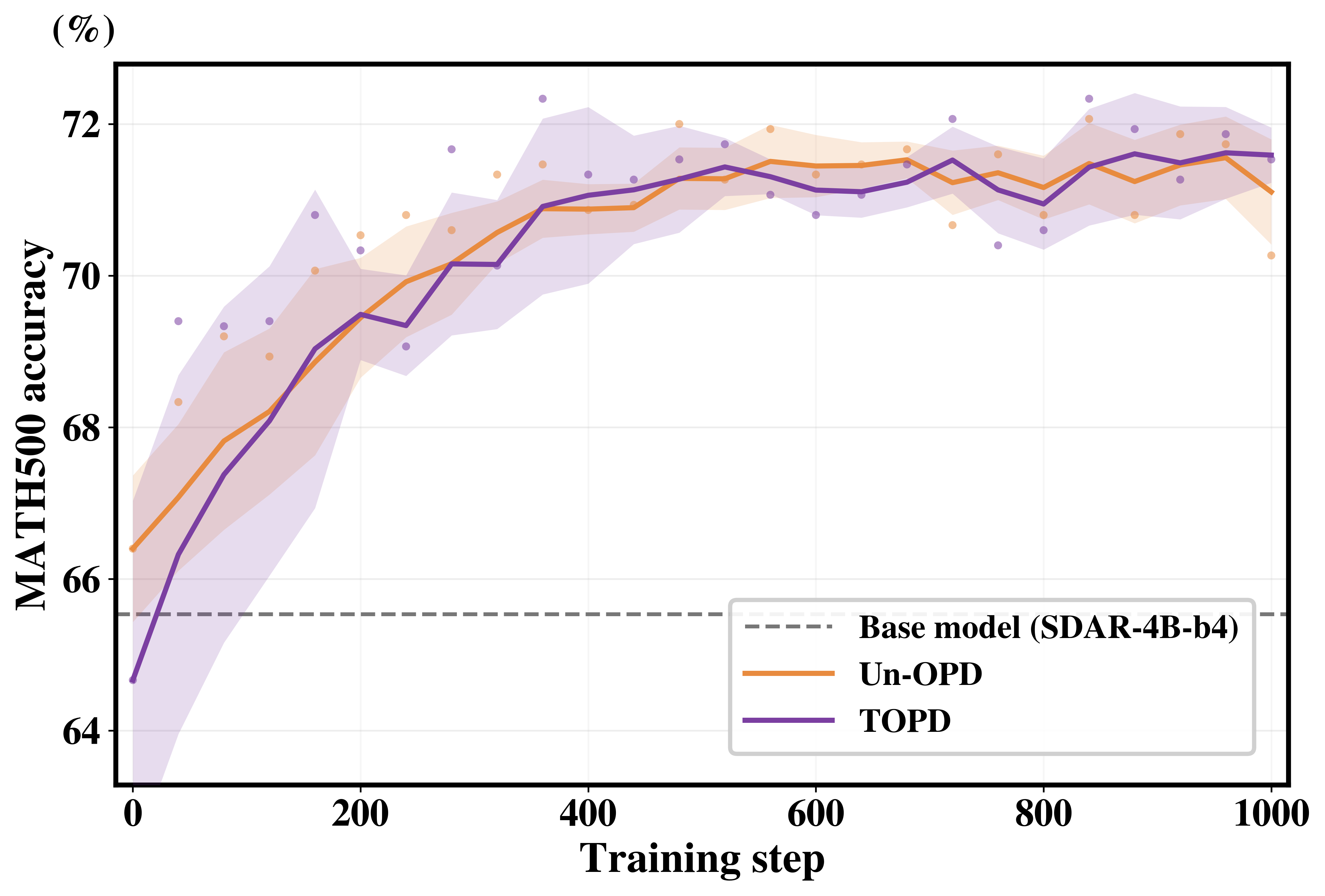}
        \caption{SDAR-4B-b4.}
        \label{fig:matched_block_training_dynamics_4b}
    \end{subfigure}
    \caption{
        \textbf{Training dynamics with matched teacher and student block sizes.}
        MATH500 accuracy under dynamic decoding with teacher and student being BDLMs with identical block sizes ($4$). Orange and purple curves correspond to Un-OPD and TOPD, respectively.
    }
    \label{fig:matched_block_training_dynamics}
\end{figure}

\subsection{Comparison with Symmetric Block Sizes}
\label{app:symmetric_block_sizes}

We further investigate training dynamics under a symmetric configuration where both the teacher and student adopt an identical block size of $4$ (Fig.~\ref{fig:matched_block_training_dynamics}).
For \texttt{SDAR-1.7B-b4}, both methods improve initially, but their trajectories markedly diverge around step 450: TOPD undergoes late-stage degradation, eroding its earlier gains, whereas \mbox{Un-OPD} stably preserves accuracy and improves further toward the end of training. In contrast, for \texttt{SDAR-4B-b4}, both methods follow comparable trajectories, plateauing around $71\%$--$72\%$ without a decisive gap.
These observations reveal that block-size symmetry alone does not eliminate the training instability in capacity-constrained students. Consequently, \mbox{Un-OPD}'s stabilizing mechanism proves crucial for weaker students, while maintaining competitive parity on larger models under symmetric regimes.

\FloatBarrier
\subsection{Extension to Larger Block Sizes}
\label{app:larger_block_sizes}

To assess whether the benefits of Un-OPD extend beyond the block sizes studied in the main experiments, we increase the student block size to $16$ and compare Un-OPD with TOPD on \texttt{SDAR-4B-b16} (\texttt{Trado-8b-b4} as teacher). Fig.~\ref{fig:larger_block_training_dynamics} reports MATH500 accuracy throughout training under both static and dynamic decoding.
Under static decoding, Un-OPD reaches approximately $72\%$ and maintains a lead over TOPD (Fig.~\ref{fig:larger_block_training_dynamics}(\subref{fig:larger_block_static})). Under dynamic decoding, it remains near $67\%$ late in training, while TOPD declines toward $61\%$ (Fig.~\ref{fig:larger_block_training_dynamics}(\subref{fig:larger_block_dynamic})). These results show that Un-OPD remains effective at a student block size of $16$, improving attainable accuracy and stability under confidence-based decoding.

\begin{figure}[H]
    \centering
    \begin{subfigure}[b]{0.49\textwidth}
        \centering
        \includegraphics[width=\linewidth]{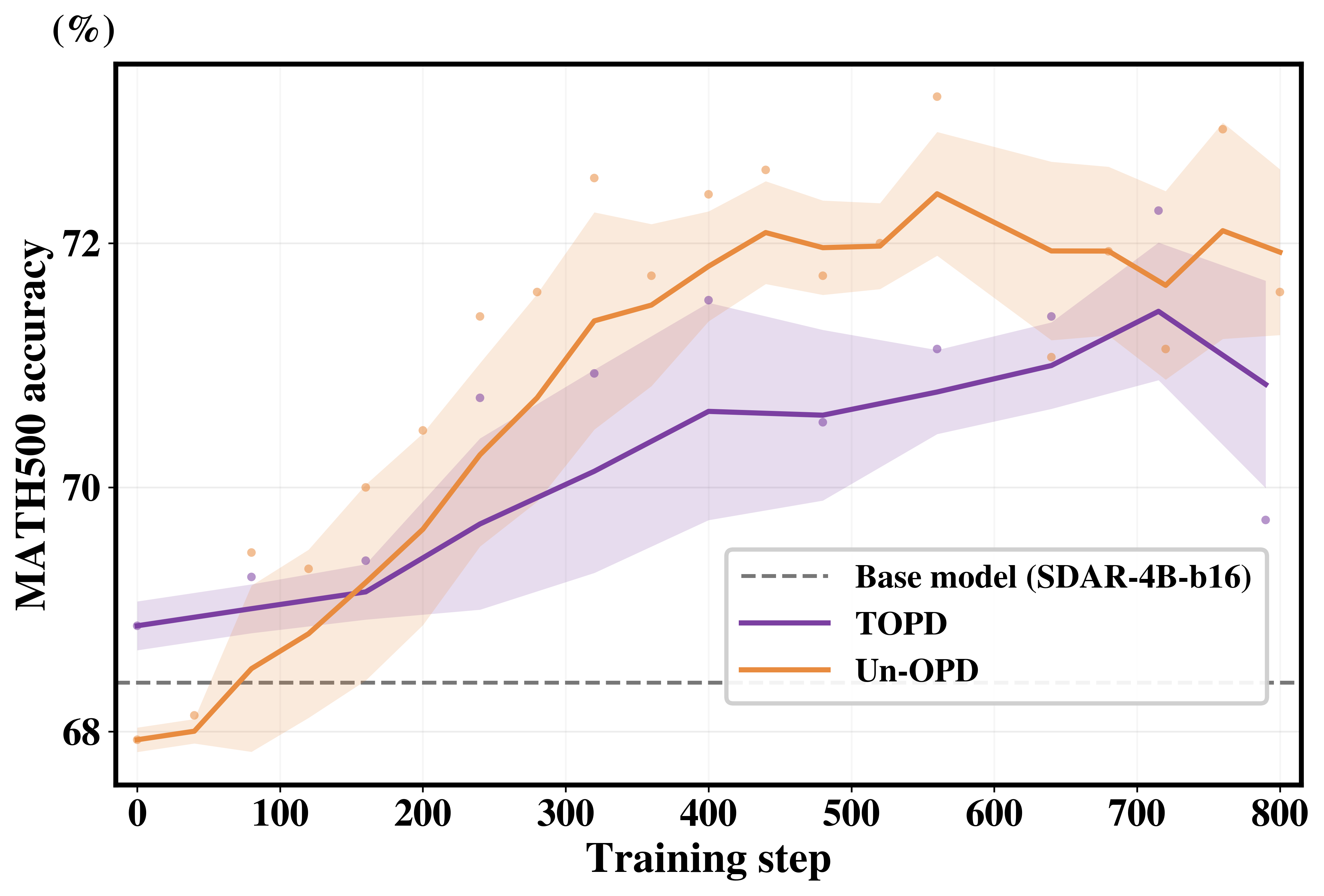}
        \caption{Static decoding.}
        \label{fig:larger_block_static}
    \end{subfigure}
    \hfill
    \begin{subfigure}[b]{0.49\textwidth}
        \centering
        \includegraphics[width=\linewidth]{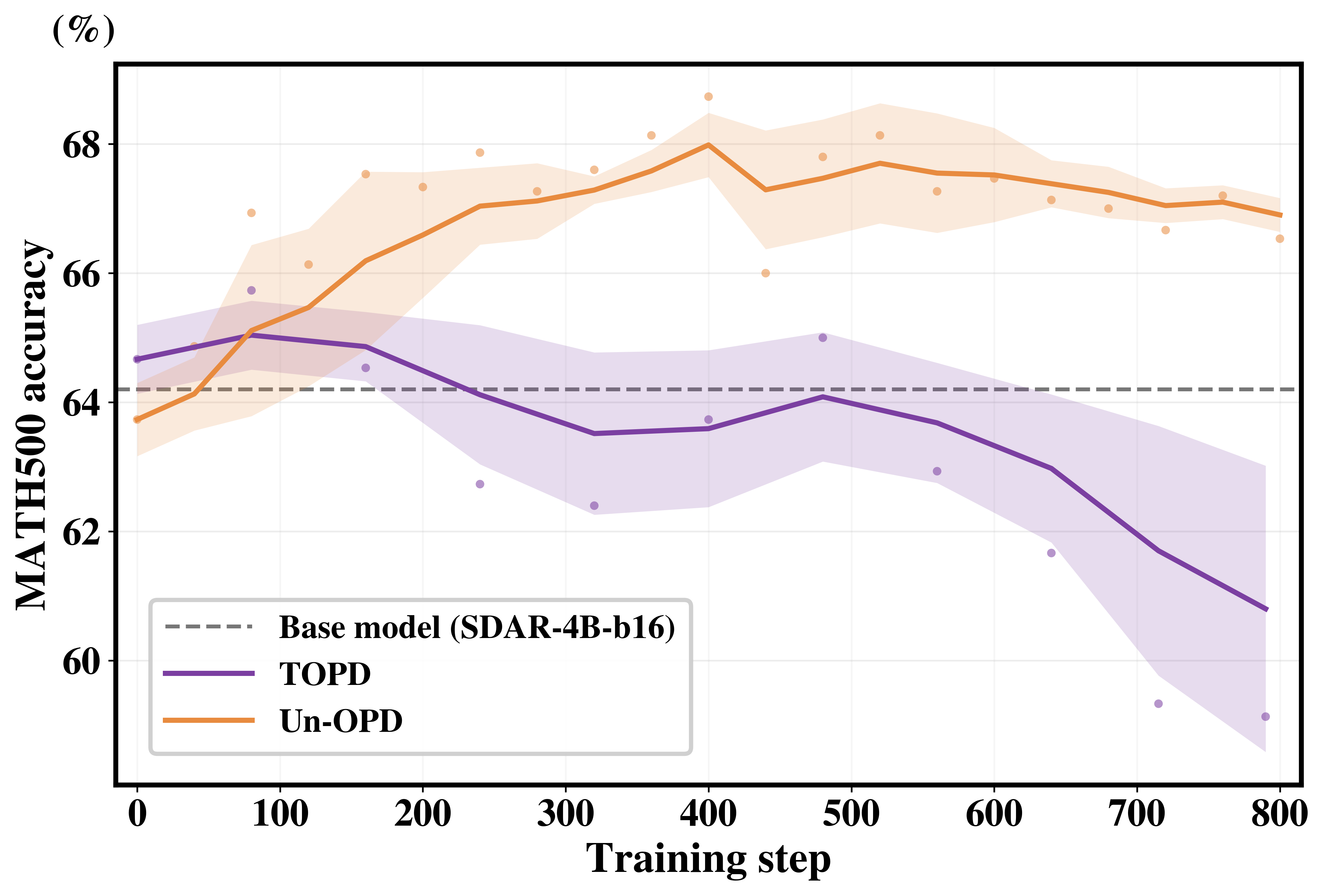}
        \caption{Dynamic decoding.}
        \label{fig:larger_block_dynamic}
    \end{subfigure}
    \caption{
        \textbf{Scaling distillation to a student block size of 16.}
    }
    \label{fig:larger_block_training_dynamics}
\end{figure}

\FloatBarrier
\subsection{Response Length Analysis}
\label{app:response_length_analysis}

We additionally track generated response length for \texttt{SDAR-1.7B} with block sizes $b\in\{4,8\}$ to examine whether training is accompanied by a sustained increase in the output-token budget. 
Fig.~\ref{fig:dynamic_response_length_1p7b} plots the average response length on MATH500 under dynamic decoding across training checkpoints. Length is measured in generated response tokens, excluding the prompt and padding.

\begin{figure}[htbp]
    \centering
    \includegraphics[width=0.76\textwidth]{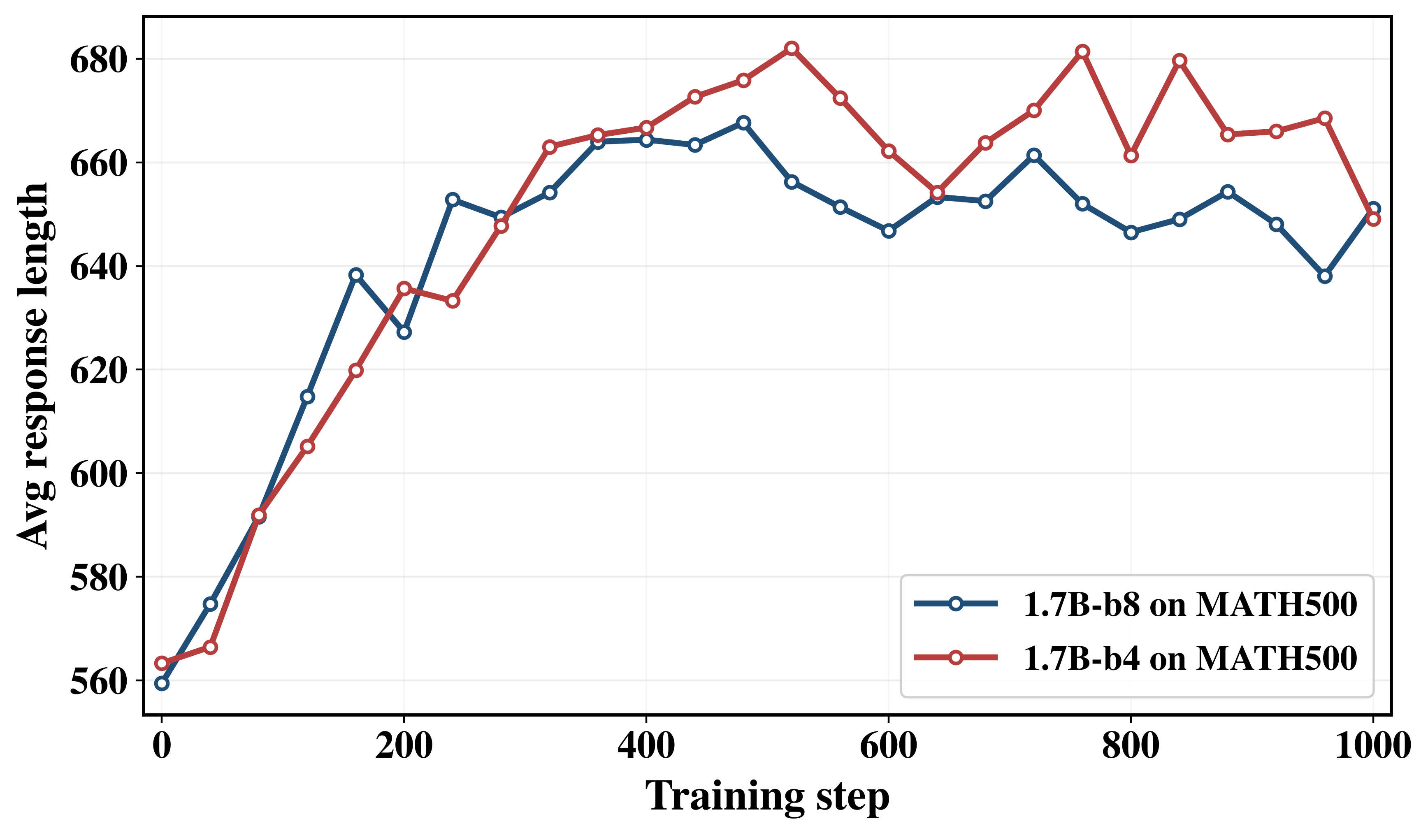}
    \caption{
        \textbf{Response length over training under dynamic decoding.}
        Average generated response length on MATH500 for SDAR-1.7B with block sizes $b=8$ (blue) and $b=4$ (red). Markers indicate the checkpoint averages.
    }
    \label{fig:dynamic_response_length_1p7b}
\end{figure}

Both trajectories rise from approximately 560 tokens at initialization to around 650--670 tokens. Thereafter, average response length fluctuates within a comparatively narrow range. 
The $b=4$ student generally produces somewhat longer responses than the $b=8$ student.
The initial increase in response length should be considered when interpreting early accuracy gains. The subsequent plateau suggests that later training is not accompanied by an uncontrolled expansion of generated responses.

\FloatBarrier

\section{Proofs}
\label{app:agreement_rkl_gradient}

We prove Proposition~\ref{prop:agreement_head_gradient} using the student's top-1 token and the conditional tail KL defined in Sec.~\ref{sec:agreement_calibrated_distillation}.

\subsection{Proof of Proposition~\ref{prop:agreement_head_gradient}}
\label{app:proof_top1_logit_asymmetry}

\textbf{Top-1 token and conditional tail distributions.}
Fix a prediction position $i$, its context, and a teacher distribution with positive token probabilities. Let $p_{\mathrm{S},i}=\operatorname{softmax}(z_{\mathrm{S},i})$ and fix the student's current top-1 token $h_i=\arg\max_v p_{\mathrm{S},i}(v)$ when evaluating the update. Following the main text, write $a_i=p_{\mathrm{S},i}(h_i)$, $b_i=p_{\mathrm{T},i}(h_i)$, and $\rho_i=b_i/a_i$. For a vocabulary with at least two tokens and finite student logits, $0<a_i,b_i<1$.
The tail consists of all vocabulary tokens other than $h_i$ at this position. Conditioning each distribution on this set gives
\[
    \bar p_{\mathrm{S},i}(v)=\frac{p_{\mathrm{S},i}(v)}{1-a_i},\qquad
    \bar p_{\mathrm{T},i}(v)=\frac{p_{\mathrm{T},i}(v)}{1-b_i},\qquad v\neq h_i.
\]
Both conditional distributions sum to one, so their reverse KL is nonnegative:
\[
    D_i^{\mathrm{tail}}
    =D_{\mathrm{KL}}(\bar p_{\mathrm{S},i}\Vert\bar p_{\mathrm{T},i})
    =\sum_{v\neq h_i}\bar p_{\mathrm{S},i}(v)
    \log\frac{\bar p_{\mathrm{S},i}(v)}{\bar p_{\mathrm{T},i}(v)}\ge0.
\]
This definition uses the student's top-1 token to partition both distributions; it does not require the teacher to have the same top-1 token.

\textbf{Reverse-KL decomposition.}
Substituting $p_{\mathrm{S},i}(v)=(1-a_i)\bar p_{\mathrm{S},i}(v)$ and $p_{\mathrm{T},i}(v)=(1-b_i)\bar p_{\mathrm{T},i}(v)$ for $v\neq h_i$ into Eq.~\ref{eq:opd_token_rkl} yields
\begin{equation}
    \mathcal{L}_i
    =a_i\log\frac{a_i}{b_i}
    +(1-a_i)\log\frac{1-a_i}{1-b_i}
    +(1-a_i)D_i^{\mathrm{tail}}.
    \label{eq:top1_logit_loss_decomposition}
\end{equation}
The first two terms together form the Bernoulli KL between the top-1 probabilities $a_i$ and $b_i$. The final term measures the discrepancy within the conditional tail distributions, weighted by the student's total tail probability $1-a_i$.

\textbf{Top-1 logit update.}
For a fixed teacher, the softmax derivative
\[
    \frac{\partial p_{\mathrm{S},i}(u)}{\partial z_{\mathrm{S},i}(v)}
    =p_{\mathrm{S},i}(u)\left[\mathbf{1}\{u=v\}-p_{\mathrm{S},i}(v)\right]
\]
gives the full-vocabulary reverse-KL gradient
\begin{equation}
    \begin{aligned}
        \frac{\partial\mathcal{L}_i}{\partial z_{\mathrm{S},i}(v)}
        &=\sum_{u\in\mathcal{V}}\left[\log\frac{p_{\mathrm{S},i}(u)}{p_{\mathrm{T},i}(u)}+1\right]
        \frac{\partial p_{\mathrm{S},i}(u)}{\partial z_{\mathrm{S},i}(v)}\\
        &=p_{\mathrm{S},i}(v)\left[\log\frac{p_{\mathrm{S},i}(v)}{p_{\mathrm{T},i}(v)}-\mathcal{L}_i\right].
    \end{aligned}
    \label{eq:agreement_rkl_gradient}
\end{equation}
Evaluating this identity at $v=h_i$ and using Eq.~\ref{eq:top1_logit_loss_decomposition}, we obtain
\begin{equation}
    \begin{aligned}
        \frac{\partial\mathcal{L}_i}{\partial z_{\mathrm{S},i}(h_i)}
        &=a_i\left[\log\frac{a_i}{b_i}-\mathcal{L}_i\right]\\
        &=-a_i(1-a_i)\left[\log\frac{b_i(1-a_i)}{a_i(1-b_i)}+D_i^{\mathrm{tail}}\right].
    \end{aligned}
\end{equation}
Multiplying by $-\eta$ gives
\[
    \Delta z_{\mathrm{S},i}(h_i)
    =\eta a_i(1-a_i)\left[\log\frac{b_i(1-a_i)}{a_i(1-b_i)}+D_i^{\mathrm{tail}}\right],
\]
which is exactly Eq.~\ref{eq:agreement_head_update}.\hfill$\square$

\textbf{Implications for the update direction.}
If $\rho_i>1$, then $b_i>a_i$ and the log-odds term is strictly positive. Since $D_i^{\mathrm{tail}}\ge0$, the top-1 logit increases. If $\rho_i<1$, the log-odds term is negative, and the top-1 logit decreases if and only if
\[
    D_i^{\mathrm{tail}}<\log\frac{a_i(1-b_i)}{b_i(1-a_i)}.
\]
Equality gives a zero update, while a larger conditional tail KL produces an upward update. When $\rho_i=1$, the update reduces to $\eta a_i(1-a_i)D_i^{\mathrm{tail}}$, which is positive whenever the conditional tail distributions differ. Thus, conditional tail mismatch reinforces upward updates at high-support positions and can weaken or reverse downward updates at low-support positions, as discussed in the main text.

The result concerns a direct gradient step on the logits at one fixed context. Changes in normalized top-1 probability and entropy also depend on updates to the other logits; training through shared model parameters additionally couples different positions. The proposition therefore characterizes a local mechanism consistent with the observed asymmetry, rather than guaranteeing a particular confidence trajectory during training.

\clearpage
\section{Case Study}
\label{app:case_study}

We compare Un-OPD and TOPD responses to the same MATH500 problem. Case 1 presents the Un-OPD response, and Case 2 presents the TOPD response. Each includes the problem and reference answer. The responses are reproduced below with formatting adjusted for typesetting.

\newtcolorbox{casebox}[1]{
    enhanced, breakable,
    colback=black!2, colframe=black!30,
    boxrule=0.8pt, arc=3mm,
    left=8pt, right=8pt, top=6pt, bottom=6pt,
    before skip=12pt, after skip=12pt,
    colbacktitle=black!30, coltitle=white,
    title={#1}, fonttitle=\bfseries,
}

\begingroup
\setlength{\parindent}{0pt}
\setlength{\parskip}{0.4em}

\begin{casebox}{Case 1: Un-OPD on MATH500 (ground truth $2k+2$)}
\textbf{Problem.} Simplify $(-k + 4) + (-2 + 3k)$.

\medskip
\textbf{Un-OPD output ($2k + 2$).}

To simplify the expression $(-k + 4) + (-2 + 3k)$, we will follow these steps:

1. Combine like terms.

First, let's rewrite the expression:
\[
(- + 4) + (- + 3k)
\]

Next, distribute the negative sign in the second term:
\[
-k + 4 + -2 + 3k
\]

Now, combine the like terms:
\begin{itemize}[leftmargin=*, nosep]
    \item Combine the $k$ terms: $-k + 3k = 2k$
    \item Combine the constant constant terms: $4 - 2 = 2$
\end{itemize}

Thus, the simplified expression is:
\[
2k + 2
\]

Therefore, the simplified form of the expression $(-k + 4) + (-2 + 3k)$ is:
\[
\boxed{2k + 2}
\]
\end{casebox}

\begin{casebox}{Case 2: TOPD on MATH500 (ground truth $2k+2$)}
\textbf{Problem.} Simplify $(-k + 4) + (-2 + 3k)$.

\medskip
\textbf{TOPD output ($4k + 2$).}

To simplify the expression $(-k + 4) + (-2 + 3k)$, we start by removing the parentheses and combining like terms.

First, distribute the negative signs in the second parentheses:
\[
(-k + 4) + (-2 + 3k) = -k + 4 - 2 + 3k
\]

Next, combine the like terms. Group the $k$ terms together and the constant terms together:
\[
k + 33 + 4 - 2
\]

Combine the constants:
\[
4 - 2 = 2
\]

Now, combine the $k$ terms:
\[
k + 3k = 4k
\]

Putting it all together, we get:
\[
4k + 2
\]

Thus, the simplified form is:
\[
\boxed{4k + 2}
\]
\end{casebox}

\endgroup

\fi

\end{document}